\documentclass{article}

\usepackage{microtype}
\usepackage{graphicx}
\usepackage{booktabs} 
\usepackage{Definitions}
\usepackage{xspace}
\usepackage{colortbl}
\usepackage[svgnames]{xcolor}
\usepackage{array,multirow,graphicx}
\usepackage{makecell}
\usepackage[utf8x]{inputenc}
\usepackage{textcomp}
\usepackage{hyperref}
\usepackage{cleveref}
\usepackage{subcaption}
\usepackage{multirow}

\newcommand{\glam}{GLaM\xspace}

\definecolor{wingreen}{rgb}{0.24,0.63,0.33}
\definecolor{glamblue}{rgb}{0,0.59,1.0}

\usepackage[accepted]{icml2022}

\icmltitlerunning{GLaM: Efficient Scaling of Language Models with Mixture-of-Experts}

\begin{document}

\twocolumn[
\icmltitle{GLaM: Efficient Scaling of Language Models with Mixture-of-Experts}



\icmlsetsymbol{equal}{*}

\begin{icmlauthorlist}
\icmlauthor{Nan Du}{equal,Google}
\icmlauthor{Yanping Huang}{equal,Google}
\icmlauthor{Andrew M. Dai}{equal,Google}
\icmlauthor{Simon Tong}{Google}
\icmlauthor{Dmitry Lepikhin}{Google}
\icmlauthor{Yuanzhong Xu}{Google}
\icmlauthor{Maxim Krikun}{Google}
\icmlauthor{Yanqi Zhou}{Google}
\icmlauthor{Adams Wei Yu}{Google}
\icmlauthor{Orhan Firat}{Google}
\icmlauthor{Barret Zoph}{Google}
\icmlauthor{Liam Fedus}{Google}
\icmlauthor{Maarten Bosma}{Google}
\icmlauthor{Zongwei Zhou}{Google}
\icmlauthor{Tao Wang}{Google}
\icmlauthor{Yu Emma Wang}{Google}
\icmlauthor{Kellie Webster}{Google}
\icmlauthor{Marie Pellat}{Google}
\icmlauthor{Kevin Robinson}{Google}
\icmlauthor{Kathleen Meier-Hellstern}{Google}
\icmlauthor{Toju Duke}{Google}
\icmlauthor{Lucas Dixon}{Google}
\icmlauthor{Kun Zhang}{Google}
\icmlauthor{Quoc V Le}{Google}
\icmlauthor{Yonghui Wu}{Google}
\icmlauthor{Zhifeng Chen}{Google}
\icmlauthor{Claire Cui}{Google}
\end{icmlauthorlist}
\icmlaffiliation{Google}{Google}

\icmlcorrespondingauthor{Nan Du, Yanping Huang, and Andrew M. Dai}{dunan@google.com, huangyp@google.com, adai@google.com}

\icmlkeywords{Machine Learning, ICML}

\vskip 0.3in
]


\printAffiliationsAndNotice{\icmlEqualContribution} 

\begin{abstract}
Scaling language models with more data, compute and parameters has driven significant progress in natural language processing. For example, thanks to scaling, GPT-3 was able to achieve strong results on in-context learning  tasks. However, training these large dense models requires significant amounts of computing resources. In this paper, we propose and develop a family of language models named \glam (\textbf{G}eneralist \textbf{La}nguage \textbf{M}odel), which uses a sparsely activated mixture-of-experts architecture to scale the model capacity while also incurring substantially less training cost compared to dense variants. The largest \glam has 1.2 trillion parameters, which is approximately 7x larger than GPT-3. It consumes only 1/3 of the energy used to train GPT-3 and requires half of the computation flops for inference, 
while still achieving better overall zero, one and few-shot performance across 29 NLP tasks. 
\end{abstract}

\section{Introduction}
\label{sec:intro}


Language models have played an important role in the progress of natural language processing (NLP) in the past decade. Variants of language models have been used to produce pretrained
word vectors~\cite{mikolov2013efficient,pennington-glove}, and
contextualized word vectors~\cite{peters2018elmo,devlin2018bert} for many NLP applications.
The shift towards scaling with more data and larger models~\cite{shazeer2017outrageously,gpipe19,kaplan2020scaling} has enabled complex natural language tasks to be performed with less labeled data.
For example, GPT-3~\cite{NEURIPS2020_gpt3} and FLAN~\cite{wei2021finetuned} demonstrated the feasibility of in-context learning for few-shot or even zero-shot generalization, meaning very few labeled examples are needed to achieve good performance on NLP applications. While being effective and performant, scaling further is becoming prohibitively expensive and consumes significant amounts of energy~\cite{patterson2021carbon}.

\begin{table}[t!]
\centering
\small
\caption{Comparison between GPT-3 and \glam. In a nutshell, \glam outperforms GPT-3 across 21 natural language understanding (NLU) benchmarks and 8 natural language generative (NLG) benchmarks in average while using about half the FLOPs per token during inference and consuming about one third the energy for training.}
\label{tab:key-comparison}
\vskip 0.1in
\begin{tabular}{l@{\hskip0.03\linewidth}c@{\hskip0.03\linewidth}c@{\hskip0.03\linewidth}c@{\hskip0.03\linewidth}l}
\toprule
& &  \textbf{GPT-3} & \textbf{\glam} & relative \\\midrule
\multirow{2}{*}{cost} &  FLOPs / token (G) &  350 & \textbf{180} & \textbf{\textcolor{wingreen}{\textminus48.6\%}}\\
& Train energy (MWh) & 1287 & \textbf{456} & \textbf{\textcolor{wingreen}{\textminus64.6\%}}\\\midrule
\multirow{3}{*}{\makecell{accuracy\\on average}} & Zero-shot & 56.9 & \textbf{62.7} & \textbf{\textcolor{wingreen}{+10.2\%}}\\
&One-shot  & 61.6 & \textbf{65.5} & \textbf{\textcolor{wingreen}{+6.3\%}}\\
&Few-shot  & 65.2 & \textbf{68.1} & \textbf{\textcolor{wingreen}{+4.4\%}}\\
\bottomrule
\end{tabular}
\end{table} 

\begin{figure*}[tb]
\centering
\renewcommand\tabcolsep{1pt}
\begin{tabular}{cccc}
\begin{subfigure}[b]{0.25\textwidth}
\includegraphics[width=\textwidth]{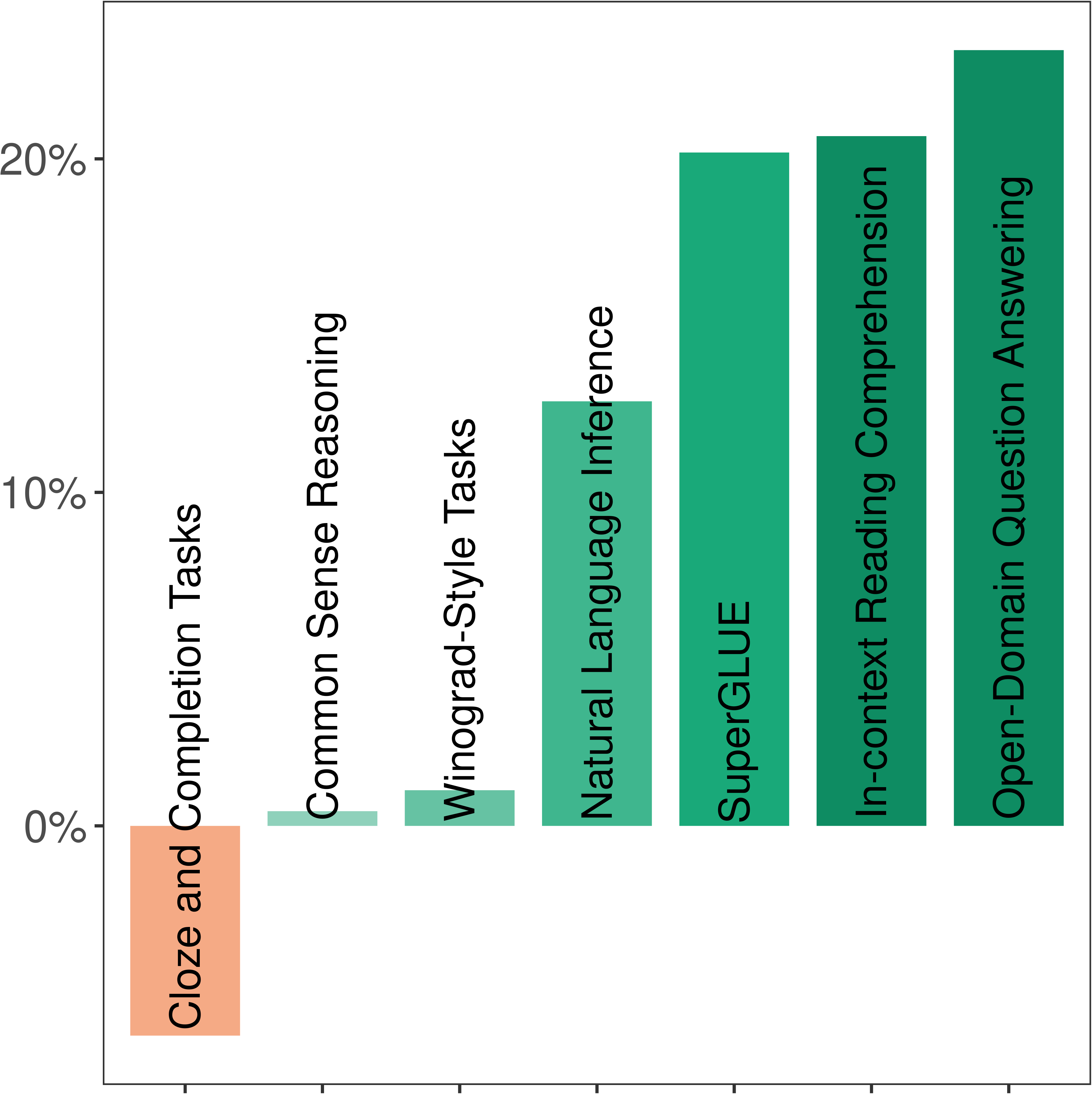}
\caption{Zero-shot}
\end{subfigure}
&
\begin{subfigure}[b]{0.25\textwidth}
\includegraphics[width=\textwidth]{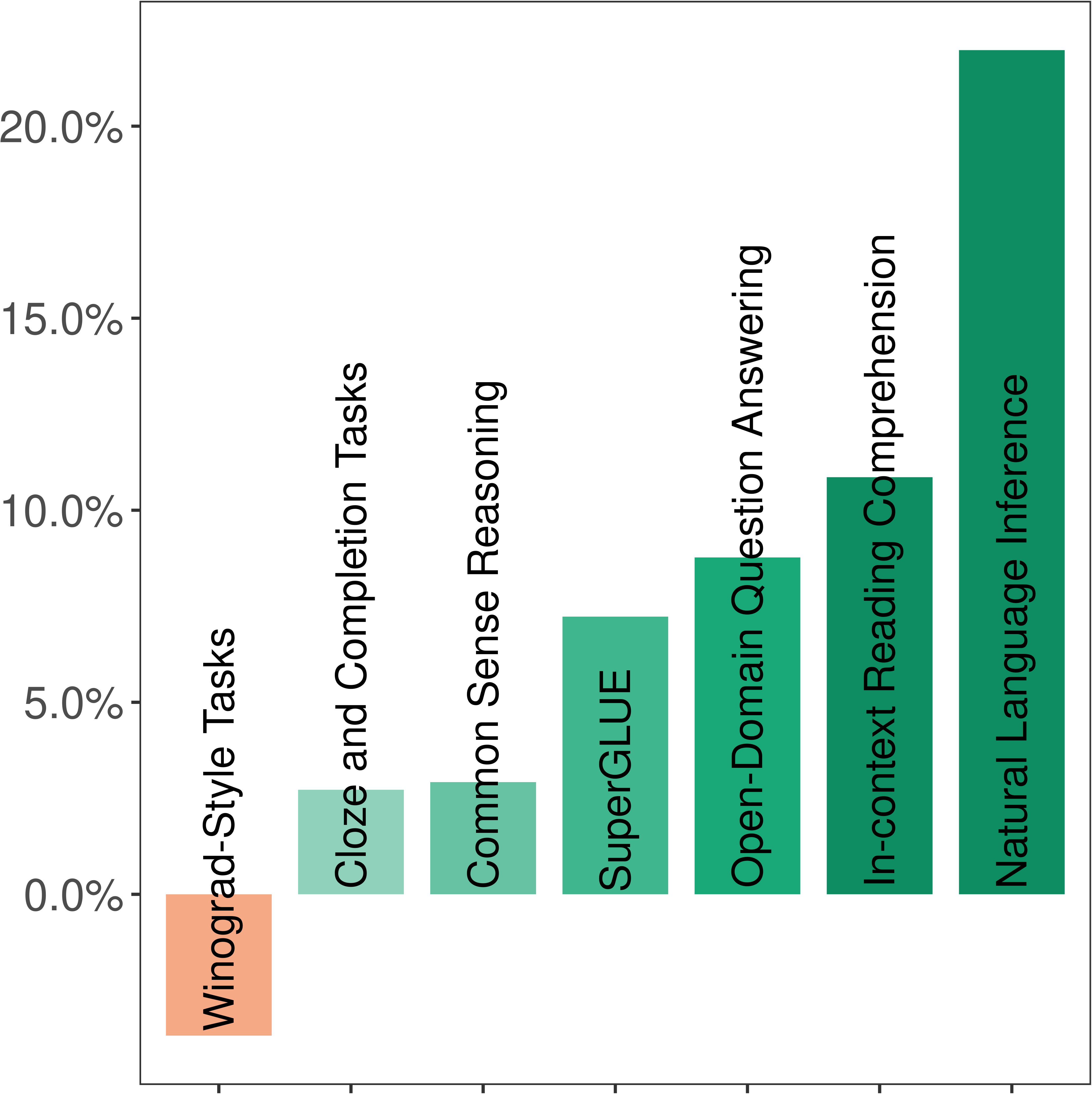}
\caption{One-shot}
\end{subfigure}
&
\begin{subfigure}[b]{0.25\textwidth}
\includegraphics[width=\textwidth]{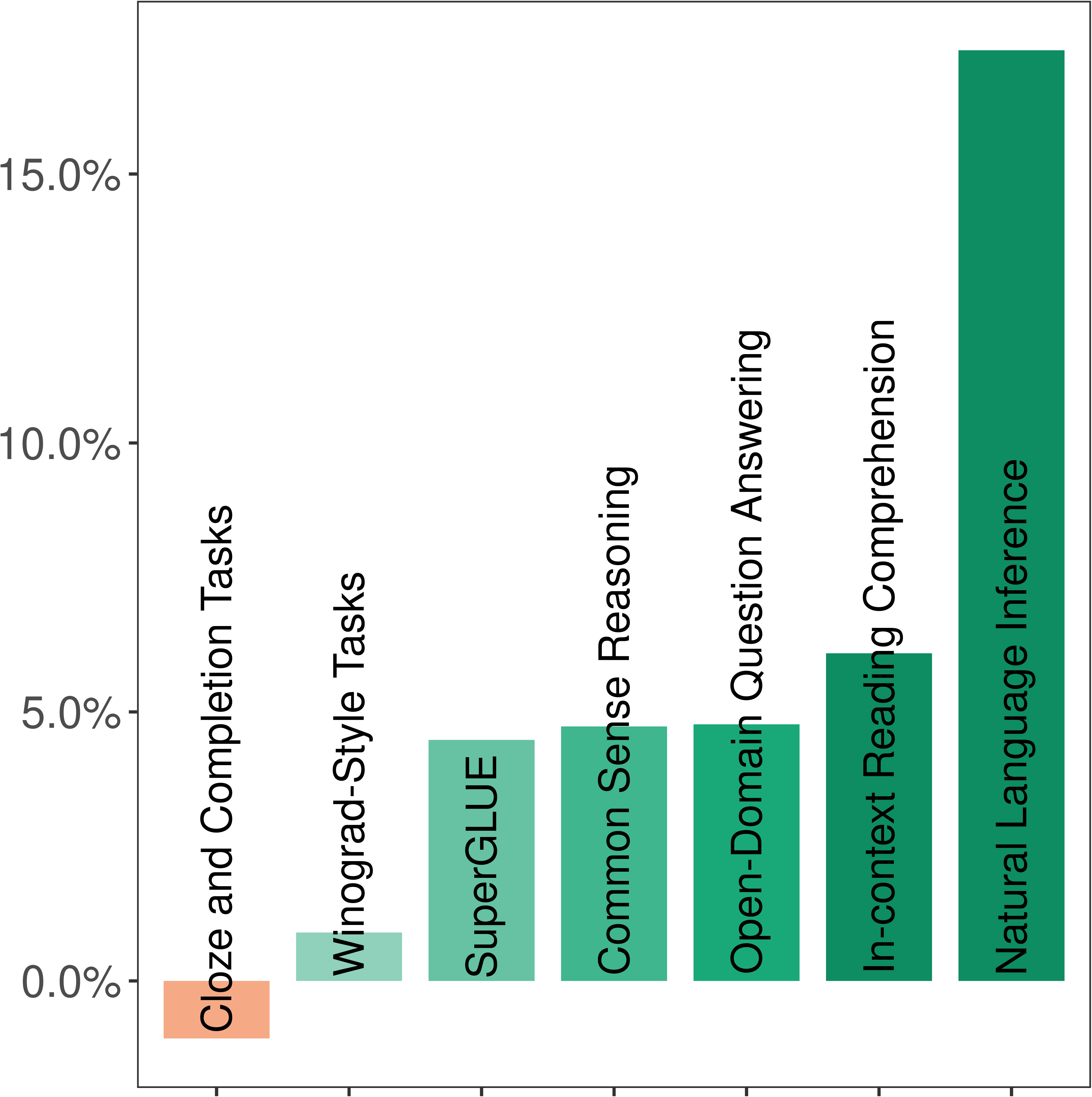}
\caption{Few-shot}
\end{subfigure}
&
\begin{subfigure}[b]{0.25\textwidth}
\includegraphics[width=\textwidth]{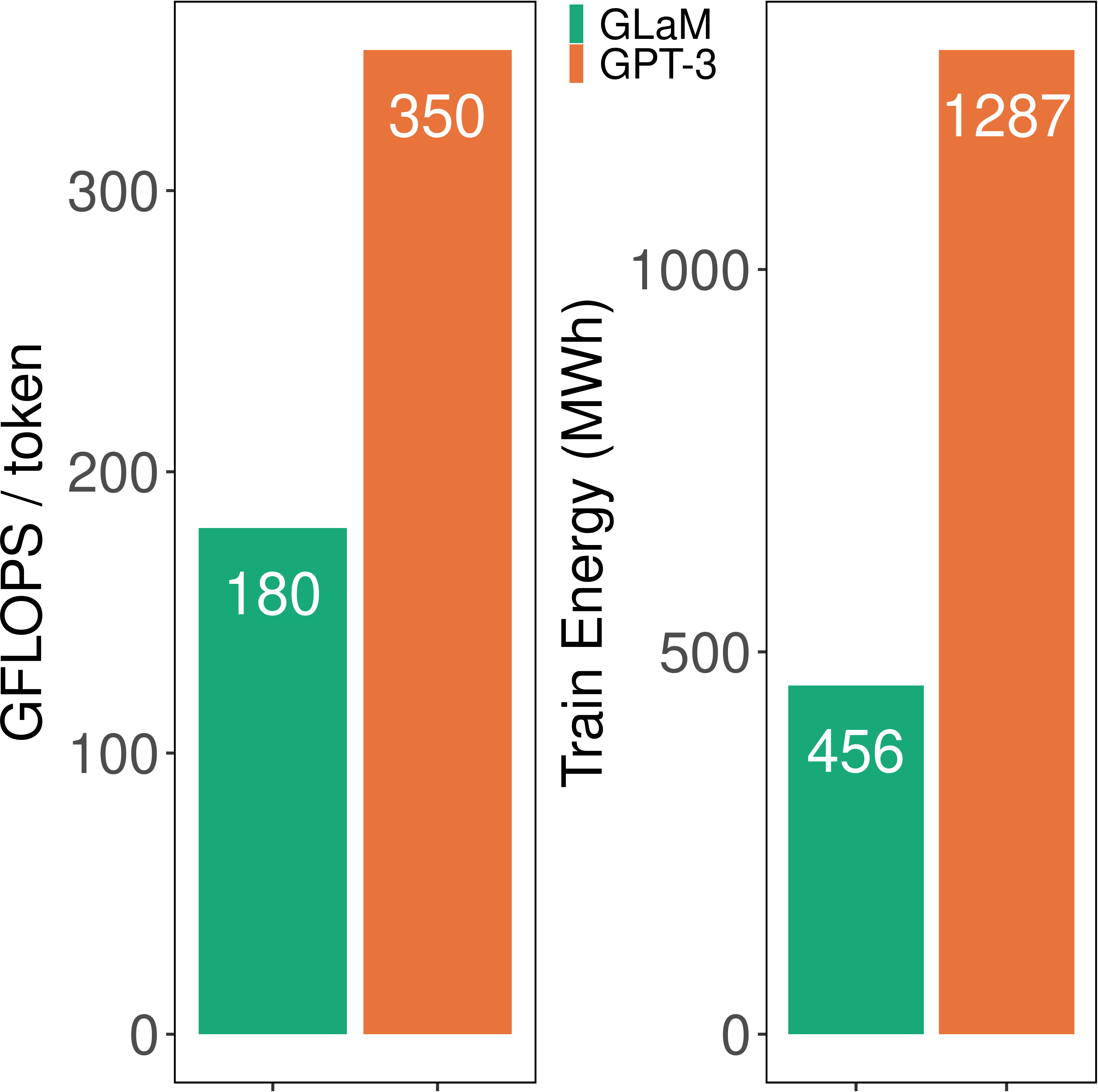}
\caption{Train and inference cost}
\end{subfigure}
\end{tabular}
\caption{An overview of the percentage change in predictive performance (higher is better) of \glam (64B/64E) versus GPT-3 (175B) in the (a) zero-shot, (b) one-shot, and (c) few-shot setting across 7 benchmark categories with 29 public tasks in total. Each bar in panel (a), (b) and (c) represents one benchmark category. Panel (d) compares the FLOPs needed per token prediction and training energy consumption.}
\label{fig:high-light-avg-by-category-01shot}
\end{figure*}

In this work, we show that a large sparsely activated network can achieve competitive results compared to state-of-the-art dense models on few-shot tasks while being more computationally efficient. We present a family of generalist language models called GLaM, that strike a balance between dense and conditional computation. The largest version of GLaM has 1.2T parameters in total with 64 experts per MoE layer~\cite{shazeer2017outrageously,lepikhin2020gshard,fedus2021switch} where each token in the input batch only activates a subnetwork of 96.6B (8\% of 1.2T) parameters. On zero, one and few-shot learning, this model compares favorably to  GPT-3 (175B), with significantly improved learning efficiency across 29 public NLP benchmarks, ranging from language completion tasks, open-domain QA tasks, to natural language inference tasks. Thanks to the sparsely activated architecture and the efficient implementation of the model parallelism algorithm, the total energy consumption during training is only one third of GPT-3's. We highlight the comparison between the largest version of \glam and GPT-3 in Table~\ref{tab:key-comparison} and Figure~\ref{fig:high-light-avg-by-category-01shot}. 

We use GLaM to study the importance of data. Our analysis shows that even for these large models, data quality should not be sacrificed for quantity if the goal is to produce a high-quality auto-regressive language model. 
More importantly, on social dimensions, our results are also the first, to our knowledge, to close the performance gap between stereotypical and anti-stereotypical examples on the WinoGender benchmark, suggesting that large, sparsely activated models may rely less on superficial statistical correlations.

Finally, although MoE-based sparse models are not yet common in the NLP community, our work shows that sparse decoder-only language models can be more performant than the dense architectures of similar compute FLOPs for the first time within the few-shot in-context learning setting at scale, suggesting 
that sparsity is one of the most promising directions to achieve high-quality NLP models while saving energy costs~\cite{patterson2021carbon}. MoE should therefore be considered as a strong candidate for future scaling. 

\section{Related Work}
\label{sec:related}

\paragraph{Language models.} Neural language models~\cite{Mikolov2010rnnlm,sutskever2011_rnnlm} have been shown to be useful for many natural language processing tasks.
Word embedding models and extensions such as word2vec~\cite{mikolov2013efficient}, GloVe~\cite{pennington-glove} and paragraph vectors~\cite{le2014distributed}
have shown good generalization to many tasks simply by transferring the embeddings. 

\paragraph{Pre-training and Fine-tuning.} The abundance of compute and data enables training increasingly large models via unsupervised pre-training. This is a natural fit for training neural networks as they exhibit remarkable scalability. Work on using recurrent models such as RNNs and LSTMs for language representation~\cite{NIPS2015_dai,kiros-skip-thought} showed that general language models could be fine-tuned to improve various language understanding tasks. More recently, models that used Transformers~\cite{vaswani2017attention} showed that larger models with self-supervision on unlabeled data could yield significant improvements on NLP tasks~\cite{devlin2018bert,yang2019xlnet,liu2019roberta,clark2020electra}. Transfer learning based on pre-training and finetuning~\cite{raffel2020exploring,houlsby2019parameterefficient} has been extensively studied and  demonstrated good performance on downstream tasks. However, a major limitation to this method is that it requires a task-specific fine-tuning. 

\paragraph{In-Context Few-shot Learning.} GPT-3~\cite{NEURIPS2020_gpt3} and related work~\cite{shoeybi2019megatron,lieber2021jurassic,wei2021finetuned} demonstrated that scaling up language models greatly improves task-agnostic, few-shot performance. These language models  are applied without any gradient updates, and only few-shot demonstrations specified purely via text interactions with the model are needed. 

\paragraph{Sparsely Gated Networks.}
Mixture-of-Experts based models have also shown significant advantages. For language modeling and machine translation, \citet{shazeer2017outrageously} showed that they could effectively use a very large number of weights while only needing to compute a small subset of the computation graph at inference time. There has also been work on scaling sparsely activated MoE architectures~\cite{hestness2017deep,shazeer2018mesh,lepikhin2020gshard, kudugunta2021beyond}. Recently, \citet{fedus2021switch} showed results with even larger 1 trillion parameter sparsely activated models (Switch-C). Although both  Switch-C and the largest \glam model have one trillion number of trainable parameters, \glam is a family of decoder-only language models, and Switch-C is an encoder-decoder based sequence to sequence model. Furthermore, 
Switch-C is mainly evaluated on fine-tuning benchmarks, \eg, SuperGlue, while \glam performs well without any need for fine-tuning in the few-shot setting shared by GPT-3 where SuperGlue is a subset.
Table~\ref{tab:various-networks} summarizes the key differences between GLaM and related models pre-trained on text corpora. 

\begin{table}[t]
{
\centering
\caption{A sample of related models~\cite{devlin2018bert,raffel2020exploring,NEURIPS2020_gpt3,lieber2021jurassic,gopher2021,shoeybi2019megatron,lepikhin2020gshard,fedus2021switch} pre-trained on text corpora. $n_{\text{params}}$ is the total number of trainable model parameters, $n_{\text{act-params}}$ is the number of activated model parameters per input token.}
	\label{tab:various-networks}
}
\vskip 0.1in
\small{
\begin{tabular}{p{2.2cm}p{3cm}p{0.75cm}p{0.9cm}}
\toprule
  Model Name & Model Type & $n_{\text{params}}$ &  $n_{\text{act-params}}$  \\
\midrule
 BERT     & Dense Encoder-only       & 340M & 340M  \\
 T5      & Dense Encoder-decoder & 13B & 13B   \\ 
 GPT-3     & Dense Decoder-only      & 175B & 175B   \\ 
 Jurassic-1 & Dense Decoder-only & 178B & 178B \\
 Gopher & Dense Decoder-only & 280B & 280B \\
 Megatron-530B  & Dense Decoder-only         & 530B & 530B  \\ 
 GShard-M4  & MoE Encoder-decoder & 600B & 1.5B \\
 Switch-C & MoE Encoder-decoder & 1.5T & 1.5B   \\
 GLaM (64B/64E) & MoE Decoder-only & 1.2T & 96.6B   \\
 \bottomrule
\end{tabular}
}

\end{table}

\section{Training Dataset}
\label{sec:data}

To train our model, we  build a high-quality dataset of $1.6$ trillion tokens that are representative of a wide range of natural language use cases. Web pages constitute the vast quantity of data in our unlabeled dataset. However, their quality ranges from professional writing to low-quality comment and forum pages. Similarly to~\citet{NEURIPS2020_gpt3}, we develop our own text quality classifier to produce a high-quality web corpus out of an original larger raw corpus. We use a feature hash based linear classifier for inference speed. This classifier is trained to classify between a collection of curated text (Wikipedia, books and a few selected websites) and other webpages. We use this classifier to estimate the content quality of a webpage. We then apply this classifier by using a Pareto distribution to sample webpages according to their score. This allows some lower-quality webpages to be included to prevent systematic biases in the classifier~\cite{NEURIPS2020_gpt3}.

\begin{table}[htb]
\centering
\small
\caption{Data and mixture weights in GLaM training set.}
\vskip 0.1in
\label{tab:data}
\begin{tabular}{lccc}
\toprule
Dataset & Tokens (B) & Weight in mixture \\
\midrule
Filtered Webpages & 143 & 0.42 \\
Wikipedia & 3 & 0.06\\
Conversations  & 174 & 0.28 \\
Forums & 247 & 0.02 \\
Books & 390 & 0.20 \\
News & 650 & 0.02 \\
\bottomrule
\end{tabular}
\end{table}
 
\begin{figure}[th]
    \centering
    \includegraphics[width=0.8\columnwidth]{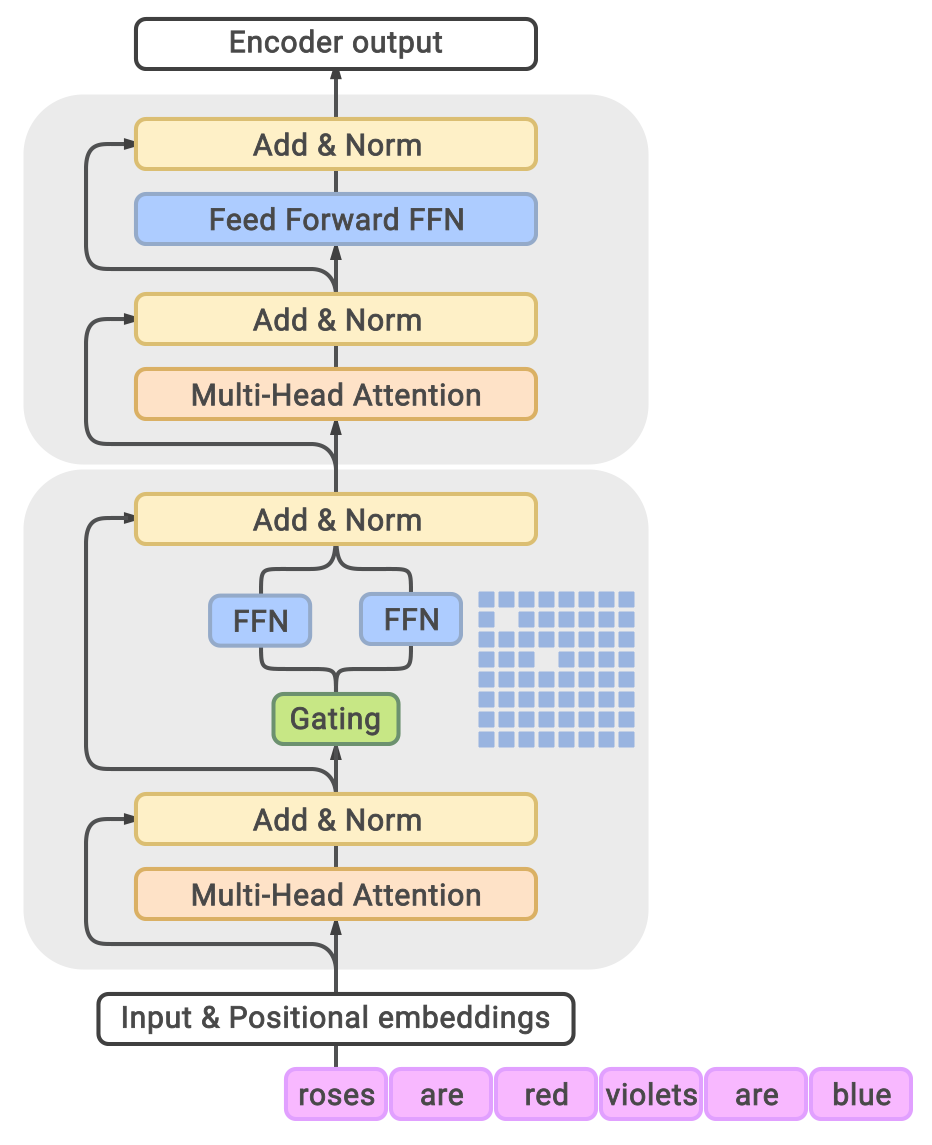}
    \caption{\glam model architecture. Each MoE layer (the bottom block) is interleaved with a Transformer layer (the upper block). For each input token, \eg, `roses', the \textit{Gating} module dynamically selects two most relevant experts out of 64, which is represented by the blue grid in the MoE layer. The weighted average of the outputs from these two experts will then be passed to the upper Transformer layer. For the next token in the input sequence, two different experts will be selected.}
    \label{fig:model}
\end{figure}
We use this process to generate a high-quality filtered subset of webpages and combine this with books, Wikipedia pages, forums and news pages and other data sources to create the final GLaM dataset. We also incorporate the data from public domain social media conversations used by~\citet{meena2020}. We set the mixture weights based on the performance of each component in a smaller model and to prevent small sources such as Wikipedia from being over-sampled. Table~\ref{tab:data} shows the details of our data component sizes and mixture weights. The mixture weights were chosen based on the performance of the component in a small model and to prevent small datasets such as Wikipedia from being over-sampled.
To check data contamination, in Section~\ref{sec:overlap} we conduct an overlap analysis between our training set and the evaluation data and find that it roughly matches that of previous work~\cite{NEURIPS2020_gpt3}.
\section{Model Architecture}
\label{sec:model}
We leverage sparsely activated Mixture-of-Experts (MoE)~\cite{shazeer2017outrageously, fedus2021switch} in \glam models. Similar to the GShard MoE Transformer~\cite{lepikhin2020gshard}, we replace the feed-forward component of every other Transformer layer with an MoE layer, as shown in Figure~\ref{fig:model}. Each MoE layer consists of a collection of independent feed-forward networks as the `experts'. A gating function then uses a softmax activation function to model a probability distribution over these experts. This distribution indicates how well each expert is able to process the incoming input. 

Even though each MoE layer has many more parameters, the experts are sparsely activated. This means that for a given input token, only a limited subset of experts is used, giving the model more capacity while limiting computation. In our architecture, the subset size is two\footnote{Using more experts will cost more compute FLOPs per prediction, pushing the network to be `denser'. Setting the number of selected experts to be two is based on the trade-off between predictive performance and the training/serving efficiency of the model.}. Each MoE layer's learnable gating network is trained to use its input to activate the best two experts for each token of an input sequence. During inference, the learned gating network dynamically picks the two best experts for each token. For an MoE layer with $E$ experts, this essentially provides a collection of $O(E^2)$ different combinations of feed-forward networks instead of one in the classic Transformer architecture, leading to much more computational flexibility.  The final learned representation of a token will be the weighted combination of the outputs from the selected experts. 

We also make additional modifications to the original Transformer architecture. We replace the standard positional embedding with per-layer relative positional bias from~\citet{daiyan2019}. In the non-MoE Transformer feed-forward sub-layers, we replace the first linear projection and the activation function with the Gated  Linear Unit~\cite{dauphin2017language, shazeer2020glu}, which computes the component-wise product of two linear transformation of the input, followed by a Gaussian Error Linear Unit~\cite{hendrycks2016bridging} activation function. 
%
\begin{table*}[tb]
    \centering
\small

    \caption{Sizes and architectures of both MoE and dense models that we have trained in our experiments. Models are grouped by the number of activated parameters per token. All trained models share the same learning hyperparameters described in Session~\ref{sec:exp_setup}.
    }
    \label{tab:setup}
    \vskip 0.1in
    \begin{tabular}{lccccccccc}
    \toprule 
    \glam Model & Type &  $n_{\text{params}}$ &  $n_{\text{act-params}}$ &  $L$ & $M$ &  $H$ &  $n_{\text{heads}}$ &  $d_{\text{head}}$ &  $E$ \\
    \midrule
    0.1B & Dense & 130M & 130M &\multirow{2}{*}{12} & \multirow{2}{*}{768} & \multirow{2}{*}{3,072} & \multirow{2}{*}{12} & \multirow{2}{*}{64} & --\\
    0.1B/64E & MoE & 1.9\text{B} & 145M & & & & & & 64\\
    \midrule
    1.7B & Dense & 1.7B & 1.700B & \multirow{5}{*}{24} & \multirow{5}{*}{2,048} & \multirow{5}{*}{8,192} & \multirow{5}{*}{16} & \multirow{5}{*}{128} & --\\
    1.7B/32E & MoE & 20B & 1.878B & & & & & & 32\\
    1.7B/64E & MoE & 27B & 1.879B & & & & & & 64\\
    1.7B/128E & MoE & 53B & 1.881B & & & & & & 128\\
    1.7B/256E & MoE & 105B & 1.886B & & & & & & 256\\
    \midrule
    8B & Dense & 8.7B & 8.7B &\multirow{2}{*}{32} & \multirow{2}{*}{4,096} & \multirow{2}{*}{16,384} & \multirow{2}{*}{32} & \multirow{2}{*}{128} & --\\ 
    8B/64E & MoE & 143B & 9.8B & & & & & & 64\\
    \midrule
    137B & Dense & 137B & 137B & 64 & 8,192 & 65,536 & 128 & 128 & --\\
    64B/64E & MoE & 1.2T & 96.6B & 64 & 8,192 & 32,768 & 128 & 128 & 64\\
    \bottomrule
    \end{tabular}

\end{table*}
We partition the weights and computation of large \glam models using the 2D sharding algorithm as described in ~\citet{xu2021gspmd}, which is described in more details in the Section~\ref{sec:gshard} of the appendix.

\section{Experiment Setup}
\label{sec:exp}
\glam is a family of dense and sparse decoder-only language models, so we first elaborate our training settings, hyperparameters, and evaluation protocol in this section.
\subsection{Training Setting}
\label{sec:exp_setup}

We train several variants of GLaM to study the behavior of MoE and dense models on the same training data. Table~\ref{tab:setup} shows the hyperparameter settings of different scale \glam models ranging from 130 million parameters to 1.2 trillion parameters. 
Here, $E$ is the number of experts in the MoE layer, $B$ is the mini-batch size, $S$ is the input sequence length, $M$ is the model and embedding dimension, $H$ is the hidden dimension of the feed-forward network, $L$ is the number of layers and $N$ is the number of total devices. Additionally,  $n_{\text{params}}$ is the total number of trainable model parameters, $n_{\text{act-params}}$ is the number of \textbf{activated} model parameters per input token, $n_{\text{heads}}$ is the number of self-attention heads, and $d_{\text{head}}$ is the hidden dimension of each attention head.
We also include the respective dense models with comparable numbers of activated parameters per-token during inference (and thus similar numbers of per-token FLOPs) as references. We adopt the notation of 
$$
\text{GLaM (Base Dense Size} / E)  \quad e.g.,~ \text{GLaM (}8\text{B}/64\text{E)}
$$
to describe different variants in the GLaM models. For example, GLaM (8B/64E) represents the architecture of an approximate 8B parameter dense model with every other layer replaced by a 64 expert MoE layer. GLaM reduces to a dense Transformer-based language model architecture when each MoE layer only has one expert. We use the notation $$\text{GLaM (Dense Size})  \quad e.g.,~ \text{\glam (137B)}$$ 
refers to a dense 137B parameter model trained with the same dataset.

\subsection{Hyperparameters and Training Procedure}
We use the same learning hyperparameters for all \glam models. More specifically, We use a maximum sequence length of $1024$ tokens, and pack each input example to have up to 1 million tokens per batch. The dropout rate is set to $0$ since the number of available tokens in the training corpus is much greater than the number of processed tokens during training. Our optimizer is Adafactor~\cite{Shazeer2018AdafactorAL} with first-moment decay $\beta_1=0$, second-moment decay $\beta_2=0.99$ with a $1 - t^{-0.8}$ decay schedule, update clipping threshold of $1.0$, and factored second-moment estimation. We keep the initial learning rate of $0.01$ for the first 10K training steps, and then decay it with inverse square root schedule $\text{lr} \langle \text{t} \rangle \propto \frac{1}{\sqrt{\text{t}}}$. On top of the standard cross-entropy loss, we add the MoE auxiliary loss as described in GShard~\cite{lepikhin2020gshard} with a $0.01$ coefficient to encourage expert load balancing so that the gating function will distribute tokens more evenly across all experts.  We use the SentencePiece~\cite{Kudo2018SentencePieceAS} subword tokenizer with a vocabulary of size of $256$K. During training, we use \textit{float32} for model weights and \textit{bfloat16} for activations. The largest GLaM 64B/64E model was trained on 1,024 Cloud TPU-V4 chips. 

Training models at the trillion parameter scale is extremely expensive even for sparsely activated models. There is little room for hyperparameter tuning. Here we share our training recipes and some implementation tricks for the \glam models.
\begin{itemize}
    \item We train smaller-scale models to convergence first. This allows us to expose potential issues in the dataset and infrastructure as early as possible.
    \item We skip weight updates for a batch if there are any \textit{NaN}s or \textit{Inf}s in the gradients~\cite{shen2019lingvo}. Note \textit{NaN/Inf} could still occur during the applying gradient step, in which case we restart from an earlier checkpoint as described below. For example, even if there is no \textit{Inf} in the existing variable or the gradient, the updated variable could still lead to \textit{Inf}.
    \item We restart from an early healthy checkpoint when encountering rare large fluctuations or even \textit{NaN/Inf} during training. Randomness of the sequentially loaded batches might help escape from previous failed states in the training after restart.
\end{itemize}

\subsection{Evaluation Setting}\label{sec:eval_setting}
\paragraph{Protocol.} To clearly demonstrate the effectiveness of \glam models, we mainly focus on evaluating the zero, one and few-shot learning protocols suggested by~\citet{radford2019language,NEURIPS2020_gpt3}. For the zero-shot learning setting, in most cases, we evaluate each example in the development set directly. For one/few-shot learning, we mainly draw random one/few examples from that task's training set as the only demonstration and context. Such a demonstration is concatenated with the evaluation example with two newlines in between, and then fed into the model.

\paragraph{Benchmarks.} To allow for an apples-to-apples comparison between GPT-3 and \glam, we choose the same suite of evaluation tasks as \citet{NEURIPS2020_gpt3}. 
But for simplicity, we exclude 7 synthetic tasks (arithmetic and word unscramble) and 6 machine translation datasets. With this exclusion, we end up with 29 datasets, which includes 8 natural language generative (NLG) tasks and 21 natural language understanding (NLU) tasks. These datasets can be further grouped into 7 categories and are listed in section~\ref{sec:benchmarks}.  

\paragraph{Natural Language Generative tasks.} We compare the language sequences decoded by the models to the ground truth in generative tasks. These tasks are TriviaQA, NQS, WebQS, SQuADv2, LAMBADA, DROP, QuAC and CoQA. The performance is measured by the accuracy of exact match (EM) and F1 score, following the standard for each task in~\citet{NEURIPS2020_gpt3}. We use beam search with a width of 4 to generate the sequences.

\paragraph{Natural Language Understanding tasks.} Most language understanding tasks require the model to select one correct answer from multiple options. All binary classification tasks are formulated into the form of selecting among two options (`Yes' or `No'). The prediction is based on the maximum log-likelihood of each option given the context $\log{P(\text{option}|\text{context})}$ normalized by the token length of each option. On a few tasks, such as ReCoRD~\cite{record2018} and COPA~\cite{COPA2012}, the non-normalized loss can yield better results and thus is adopted. Except for MultiRC~\cite{multirc2018} where the $\text{F1}$ metric over the set of answer options (referred to as $\text{F1}_a$) is reported, the prediction accuracy metric is used for all the other tasks.
We use the average of the scores reported in all datasets to report the overall few-shot performance of models on both NLG and NLU tasks.  Both Accuracy (EM) and F1 scores have been normalized to lie between 0 and 100. On TriviaQA, we also report the testing server score of our one-shot submission. 

\section{Results}

We conduct extensive evaluation on the whole family of \glam models, to show the advantages of sparsely activated models in language modeling and their scaling trends. We also quantitatively inspect the effectiveness of data quality for language model training.

\subsection{Comparison between MoE and Dense Models}\label{sec:moe_vs_dense}
As previously presented in Table~\ref{tab:key-comparison}, \glam (64B/64E) has competitive performance compared to GPT-3 (175B) for zero, one and few-shot learning. Figure~\ref{fig:high-light-avg-by-category-01shot} compares the performance for each category of tasks. In total, \glam (64B/64E) outperforms GPT-3 in 6 out of 7 categories on average, indicating the performance gain is consistent. For more details on each individual task, see Table~\ref{tab:main-results}. We include results on the much larger and computationally demanding Megatron-NLG and Gopher for reference.
More importantly, as shown in Table~\ref{tab:setup}, \glam (64B/64E) activates roughly 96.6B parameters per token during inference, which requires only half of the compute FLOPs needed by GPT-3 given the same input.


We highlight one particular challenging open-domain question answer task: \emph{TriviaQA}. In open-domain question answer tasks, the model is required to directly answer a given query without access to any additional context.
\citet{NEURIPS2020_gpt3} show that the few-shot performance of TriviaQA is able to grow smoothly with model size, indicating a language model is able to absorb knowledge using its model capacity.
As shown in Table~\ref{tab:triviaqa}, \glam (64B/64E) is better than the dense model and outperforms the previous finetuned state-of-the-art (SOTA) on this dataset in the open-domain setting. 
Our one-shot result exceeds the previous finetuned SOTA~\cite{yu2021kgfid} where additional knowledge graph information is infused by 8.6\%, and outperforms the few-shot GPT-3 on the testing server by 5.3\%.
This suggests that the additional capacity of \glam plays a crucial role in the performance gain even though the $n_{\text{act-params}}$ of \glam (64B/64E) is only half of that in GPT-3. Comparing to Switch-C, even though both models have similar total number of parameters, \glam (64B/64E) uses much larger experts (beyond one TPU core) than Switch-C. Therefore, GLaM’s one-shot performance on TriviaQA is also better than the fine-tuned results of Switch-C in the open-domain setting. 
\begin{table}[tb]
    \centering
    \small
        \caption{\glam (64B/64E) one-shot performance significantly outperforms prior SOTAs for open domain settings in the wiki split.}
    \label{tab:triviaqa}
    \vskip 0.1in
    \begin{tabular}{lc}
        \toprule
        Model & \makecell{TriviaQA \\(Open-Domain)} \\
        \midrule
        \makecell[l]{KG-FiD (large)~\cite{yu2021kgfid} \\({\scriptsize finetuned, test})} & 69.8\\
        Switch-C ({\scriptsize finetuned, dev}) & 47.5\\
        GPT-3 One-shot ({\scriptsize dev}) & 68.0\\
        GPT-3 64-shot ({\scriptsize test}) & 71.2\\
        \glam One-shot ({\scriptsize test})& 75.0\\ 
        \glam One-shot ({\scriptsize dev})& \textbf{75.8}\\
        \bottomrule
    \end{tabular}

\end{table}
Finally, we report zero, one and few-shot evaluation mainly on the development set for all tasks in Tables \ref{tab:main-results}, \ref{tab:0shot}, \ref{tab:1shot} and \ref{tab:kshot} of the appendix.


\begin{figure*}[tb]
\centering
    \renewcommand\tabcolsep{1pt}
\begin{tabular}{cccc}
\begin{subfigure}[b]{0.25\textwidth}
\includegraphics[width=\textwidth]{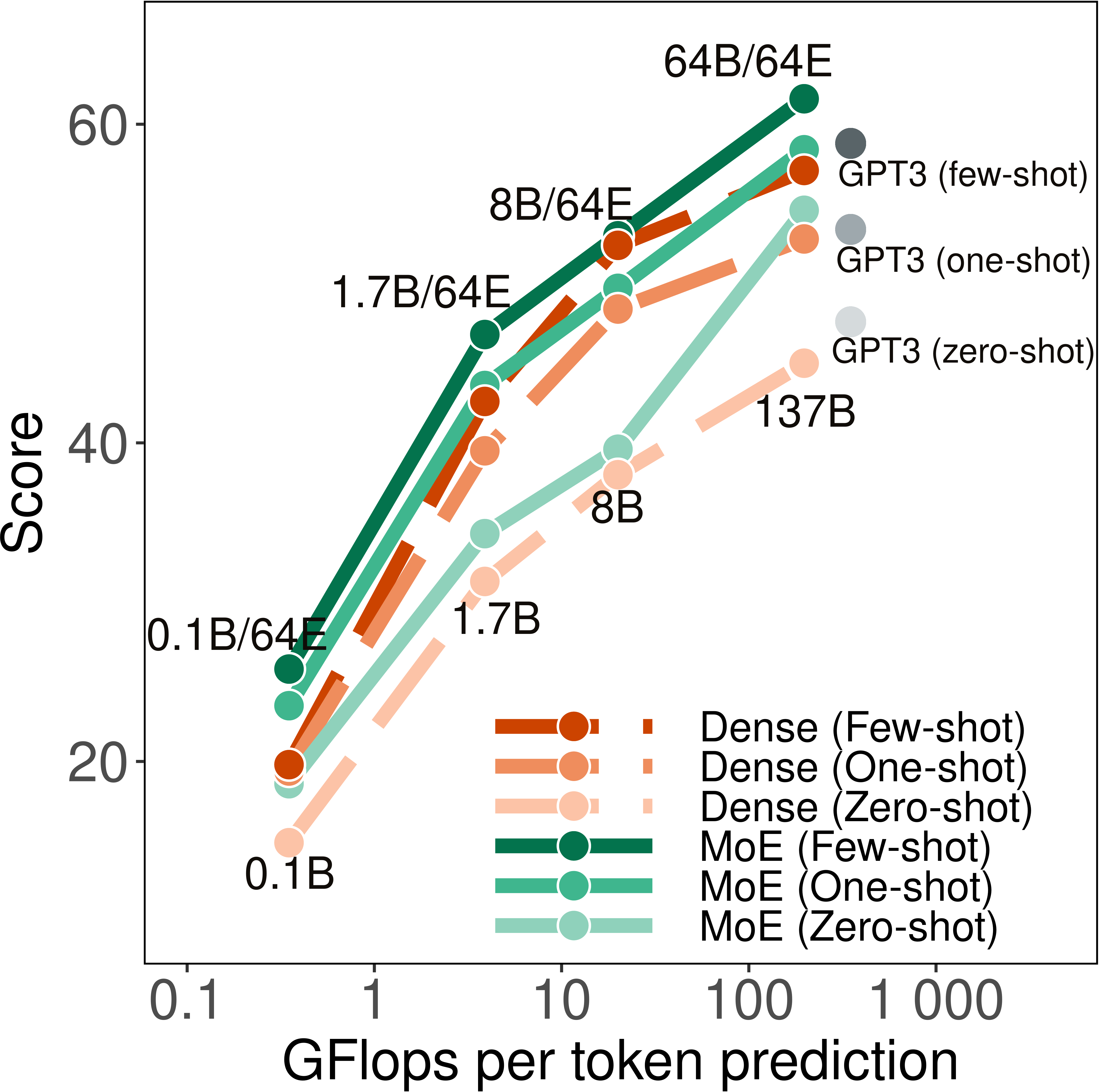}
\caption{Scaling (NLG)}
\end{subfigure}
&
\begin{subfigure}[b]{0.25\textwidth}
\includegraphics[width=\textwidth]{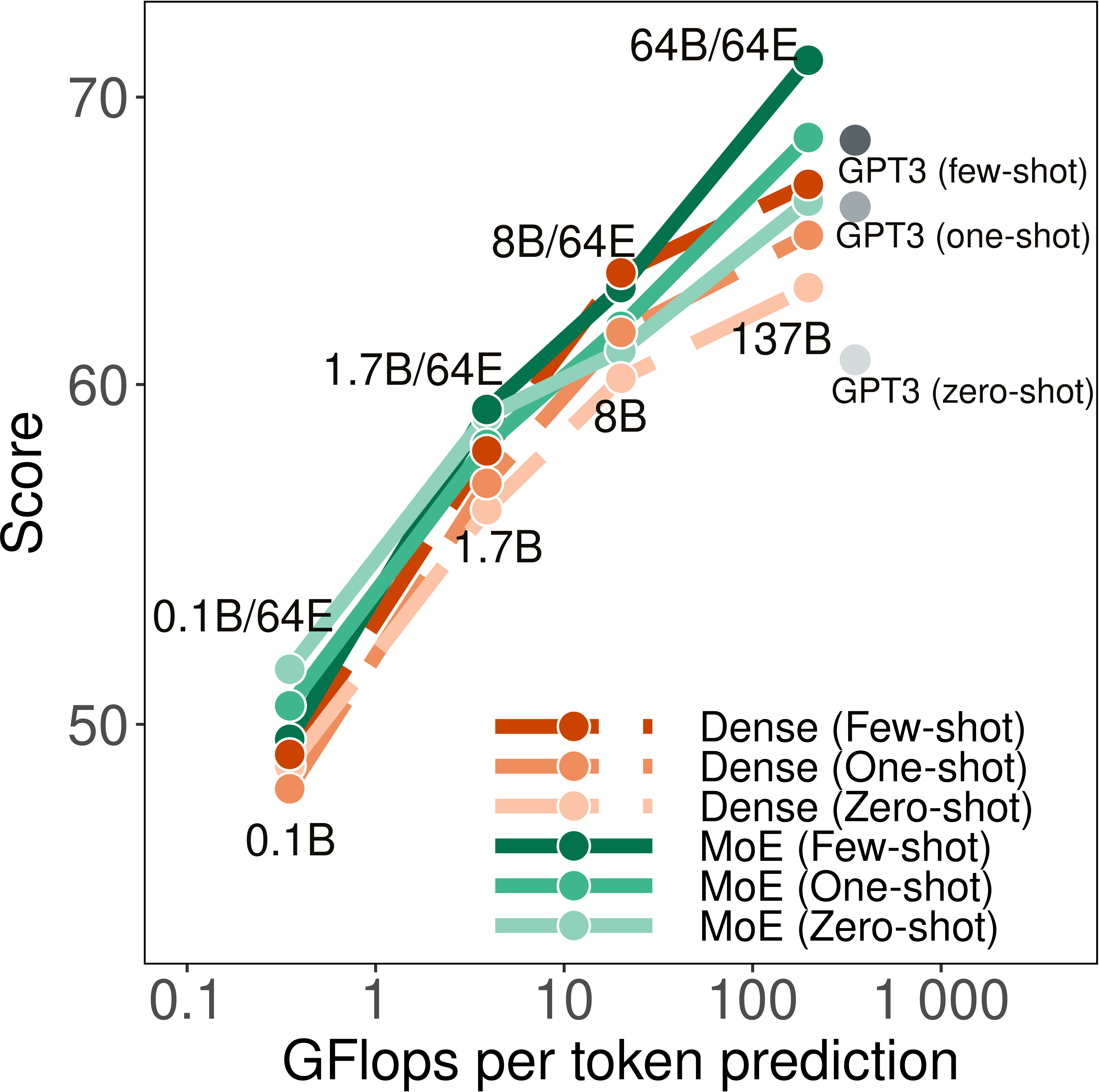}
\caption{Scaling (NLU)}
\end{subfigure}
&
\begin{subfigure}[b]{0.25\textwidth}
\includegraphics[width=\textwidth]{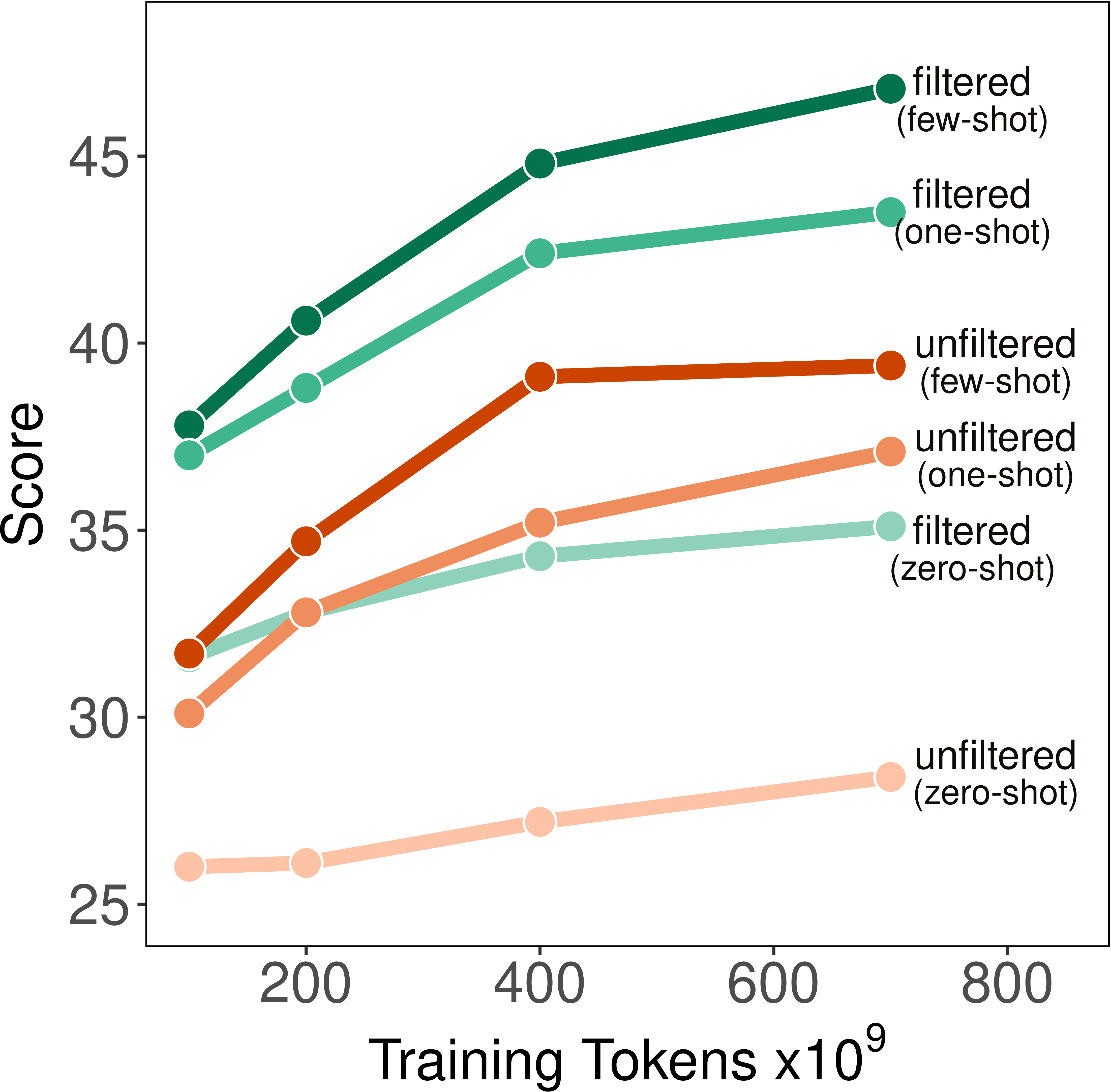}
\caption{Data filtering (NLG)}
\end{subfigure}
&
\begin{subfigure}[b]{0.25\textwidth}
\includegraphics[width=\textwidth]{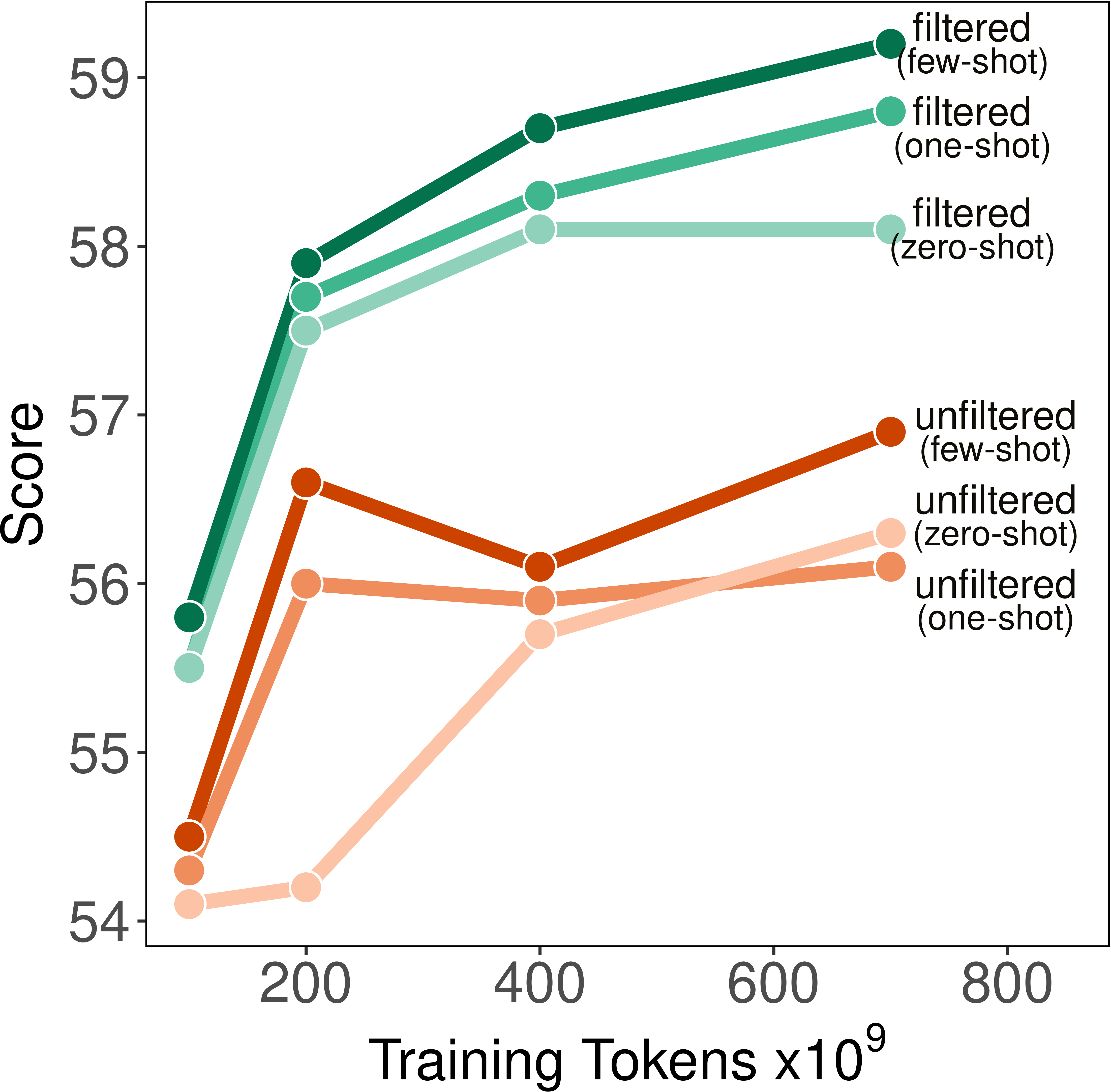}
\caption{Data filtering (NLU)}
\end{subfigure}
\end{tabular}
\caption{Average zero, one and few-shot performance of \glam MoE models versus \glam dense models for similar effective FLOPs per token over the 8 NLG tasks (a) and 21 NLU tasks (b). Comparison of model performance with filtered and unfiltered training data using \glam(1.7B/64E). Filtered data improves results significantly over unfiltered data for both (c) NLG and (d) NLU tasks across zero, one and few-shot settings.}
\label{fig:scale-size-data-efficiency}
\end{figure*}

\subsection{Effect of Data Quality}
\label{sec:data-quality}
We study the impact of data quality on the few-shot performance of downstream tasks. We use a modest-size GLaM model (1.7B/64E) to show the effectiveness of filtering text on model quality. We train models with the same hyperparameters on two datasets. One is the original dataset described in Section~\ref{sec:data} and the second consists of the dataset with the filtered webpages replaced with the unfiltered webpages. The mixing proportions are fixed as given in Table~\ref{tab:data}. The filtered webpages consist of 143B tokens whereas the unfiltered webpages consist of around 7T tokens.

Figure~\ref{fig:scale-size-data-efficiency} (c) and (d) show that the model trained on filtered data performs consistently better on both NLG and NLU tasks. In particular, the effect of filtering is bigger on NLG than that on NLU. Perhaps this is because NLG often requires generating high-quality language and filtered pretraining corpora is crucial to the generation capability of language models.
Our study highlights the fact that the quality of the pretrained data also plays a critical role, specifically, in the performance of downstream tasks.

\subsection{Scaling Studies}\label{sec:model_scaling}
Scaling up dense language models generally involves making the models deeper by adding more layers, and wider by increasing the embedding dimension of token representations. This process increases the total number of parameters $n_{\text{params}}$ of the model. For each prediction on a given input example, these models are `dense' in that all $n_{\text{params}}$ parameters will be activated, i.e., $n_{\text{params}} = n_{\text{act-params}}$ in Table~\ref{tab:setup}. Therefore, the effective FLOPs per prediction increases linearly with the model size $n_{\text{params}}$. While the increased FLOPs may lead to boosted predictive performance, it also raises the overall cost per prediction.

In contrast, \glam MoE models are \textbf{sparsely activated} in that only a small fraction of the total $n_{\text{params}}$ parameters will be activated for each prediction where $n_{\text{params}} \gg n_{\text{act-params}}$. Therefore,  
\glam MoE models can scale by also growing the size or number of experts in the MoE layer.






As shown in Figure~\ref{fig:scale-size-data-efficiency}(a), the average zero, one and few-shot performance across the generative tasks scales well with the effective FLOPs per prediction which is in turn determined by $n_{\text{act-params}}$. We also find that \glam MoE models perform consistently better than \glam dense models for similar effective FLOPs per token. For language understanding tasks shown in Figure~\ref{fig:scale-size-data-efficiency}(b), the performance gain of \glam MoE models has a similar scaling trend to that of the generative tasks. We observe that both MoE and dense models perform similarly at smaller scales but MoE models outperform at larger scales. We also show experiments with scaling the number of experts in Section~\ref{sec:scaling_number_experts} where we observe that, for a fixed budget of computation per prediction, adding more experts generally leads to better predictive performance.
\subsection{Efficiency of \glam}
Existing large dense language models usually require tremendous amounts of computation resources for training and serving~\cite{patterson2021carbon}. They also need to consume massive amounts of pretraining data. 
We investigate the data and compute efficiency of the proposed \glam models.

\begin{figure*}[tb]
\centering
\renewcommand\tabcolsep{2pt}
\begin{tabular}{cccc}
\begin{subfigure}[b]{0.24\textwidth}
\includegraphics[width=\textwidth]{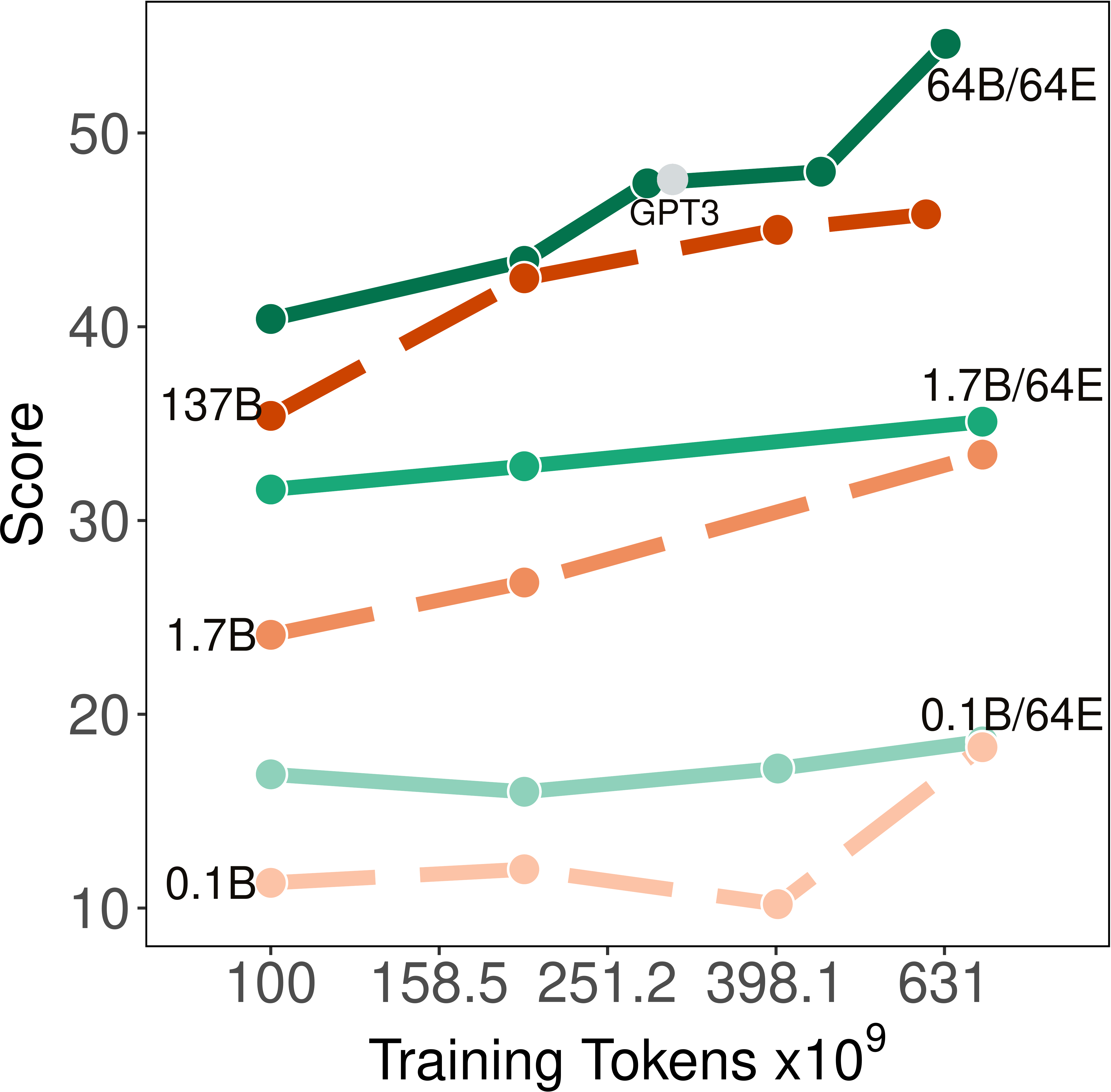}
\caption{Zero-shot (NLG)}
\end{subfigure}
&
\begin{subfigure}[b]{0.24\textwidth}
\includegraphics[width=\textwidth]{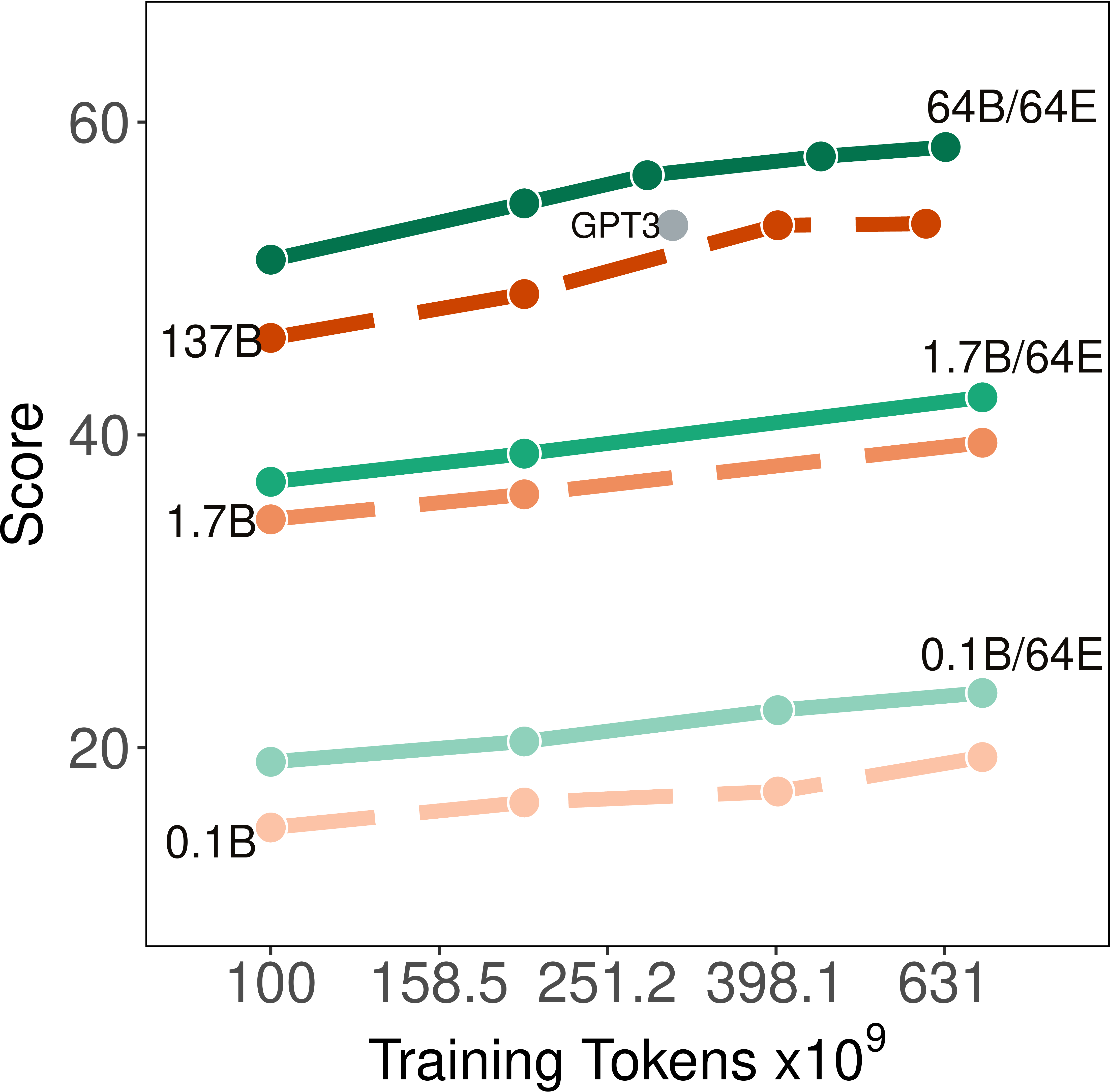}
\caption{One-shot (NLG)}
\end{subfigure}
&
\begin{subfigure}[b]{0.24\textwidth}
\includegraphics[width=\textwidth]{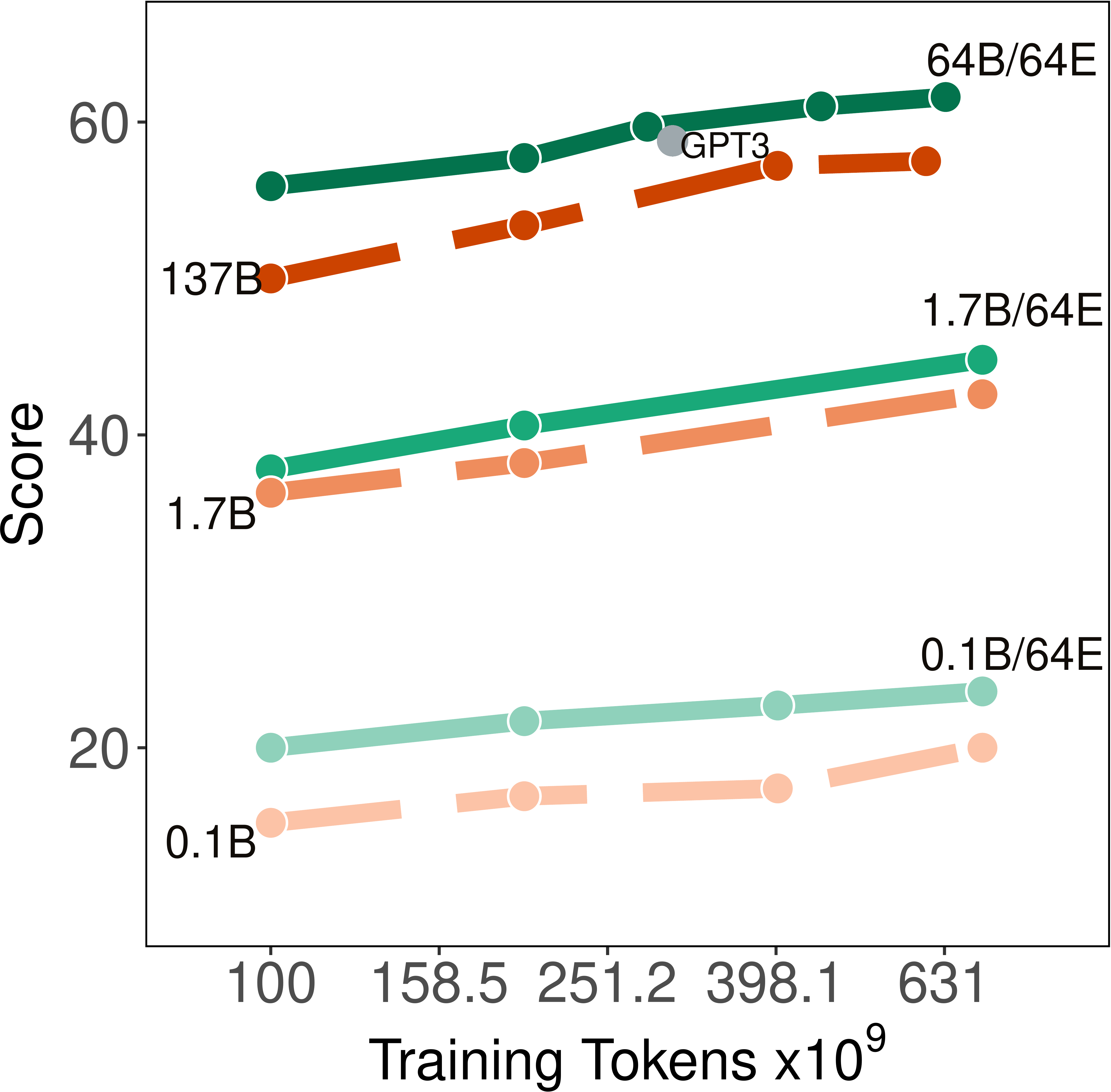}
\caption{Few-shot (NLG)}
\end{subfigure}
&
\begin{subfigure}[b]{0.24\textwidth}
\includegraphics[width=\textwidth]{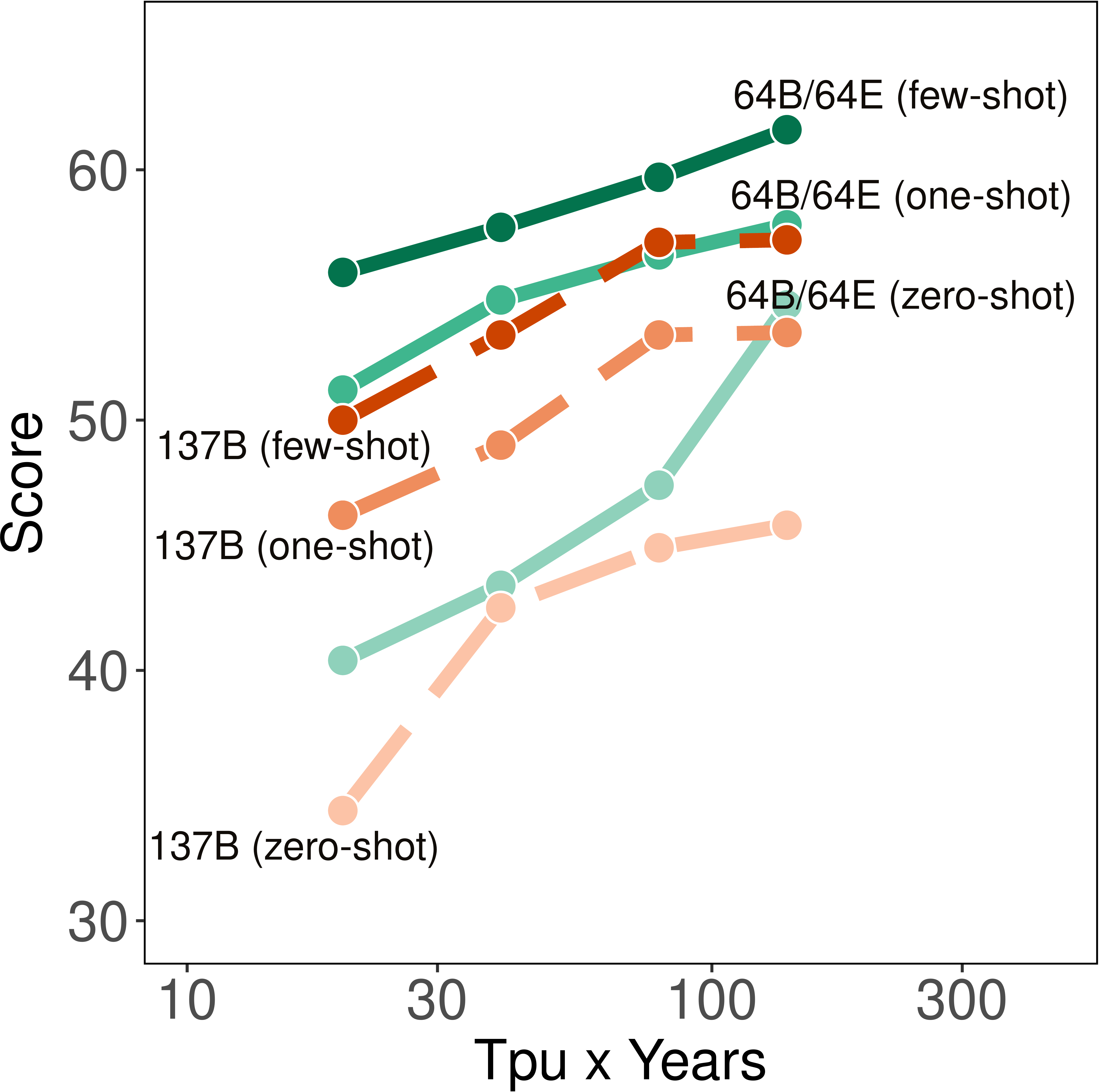}
\caption{Scaling in TPU years (NLG)}
\end{subfigure}
\\
\begin{subfigure}[b]{0.24\textwidth}
\includegraphics[width=\textwidth]{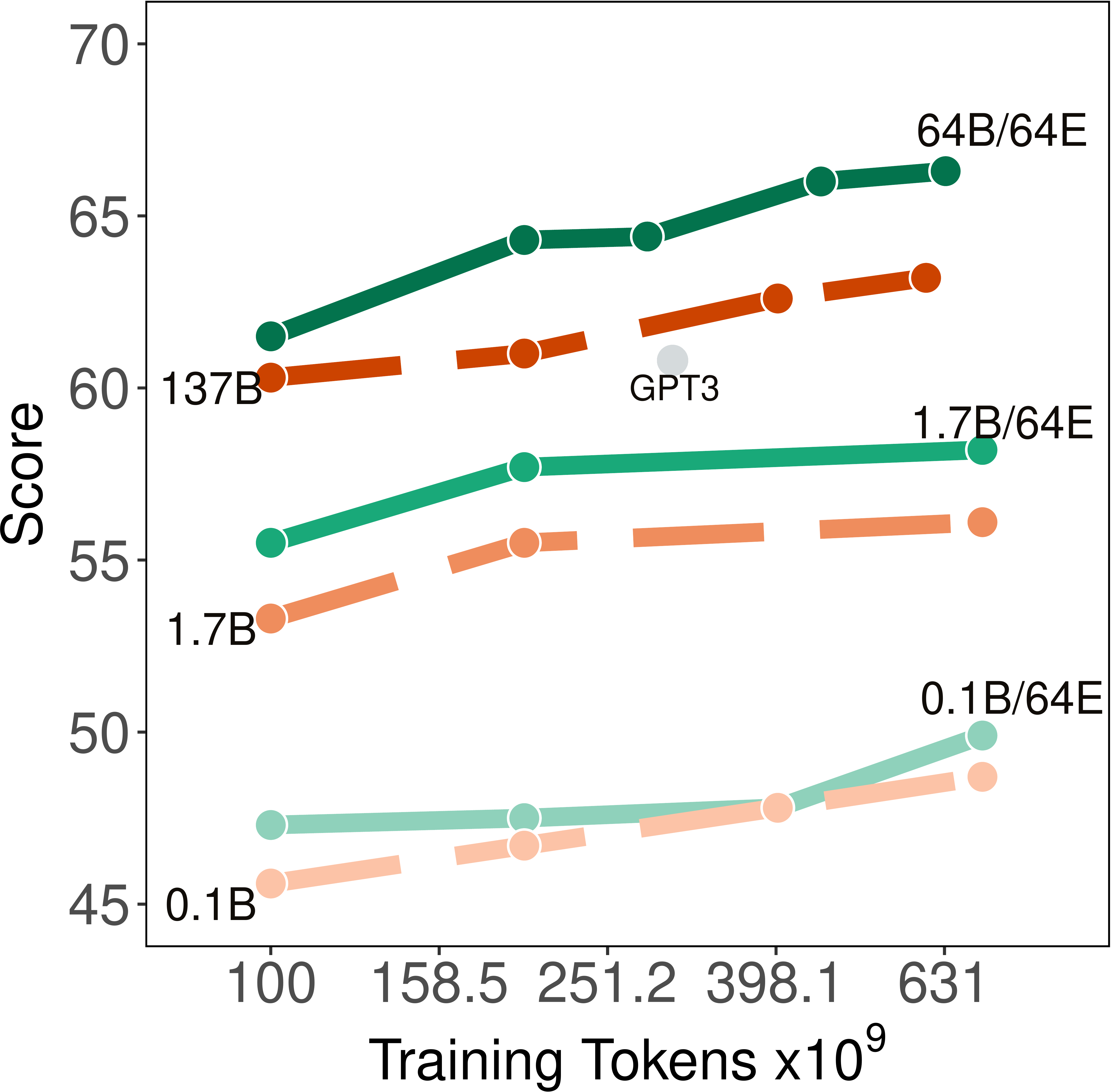}
\caption{Zero-shot (NLU)}
\end{subfigure}
&
\begin{subfigure}[b]{0.24\textwidth}
\includegraphics[width=\textwidth]{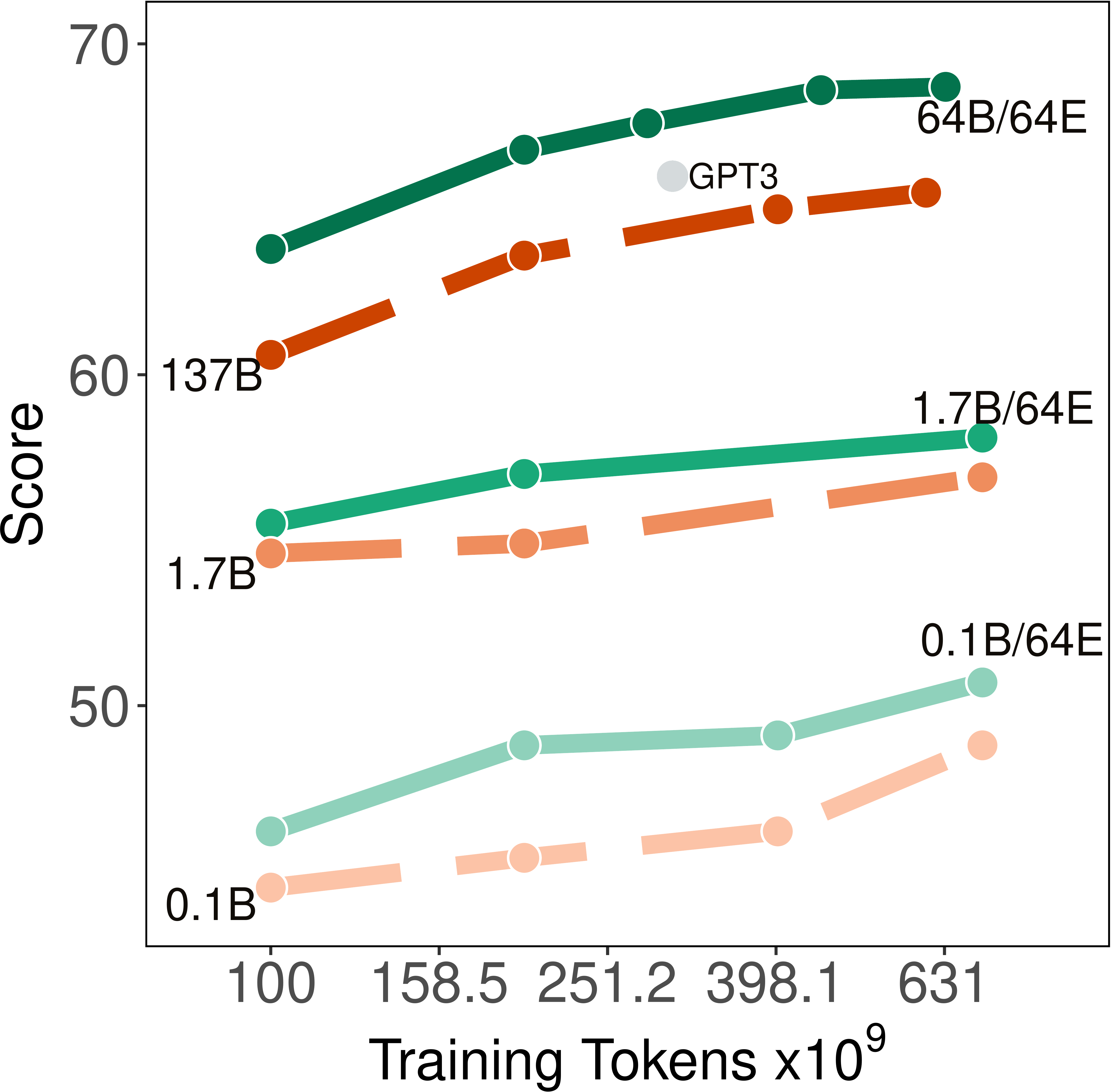}
\caption{One-shot (NLU)}
\end{subfigure}
&
\begin{subfigure}[b]{0.24\textwidth}
\includegraphics[width=\textwidth]{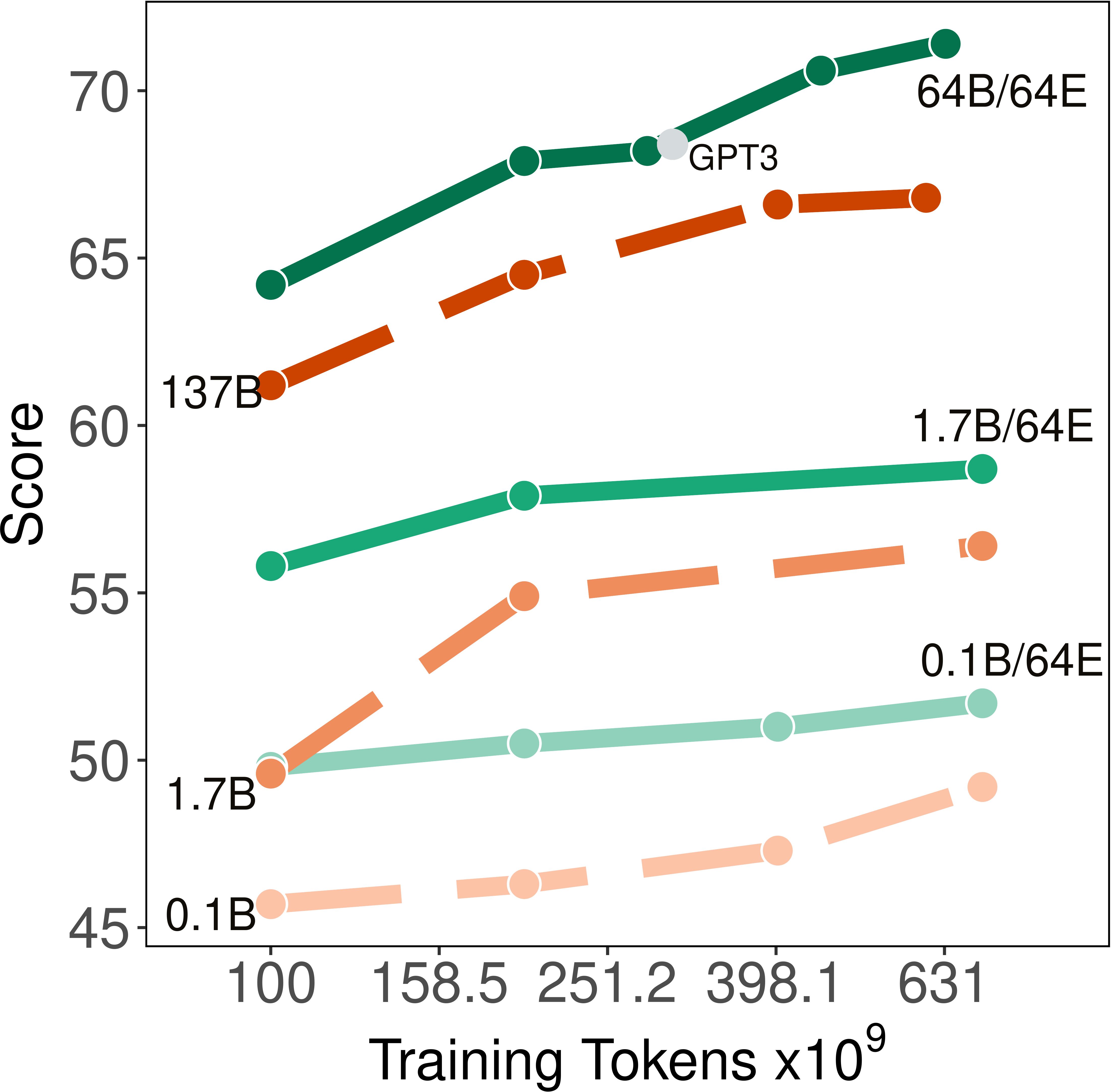}
\caption{Few-shot (NLU)}
\end{subfigure}
&
\begin{subfigure}[b]{0.24\textwidth}
\includegraphics[width=\textwidth]{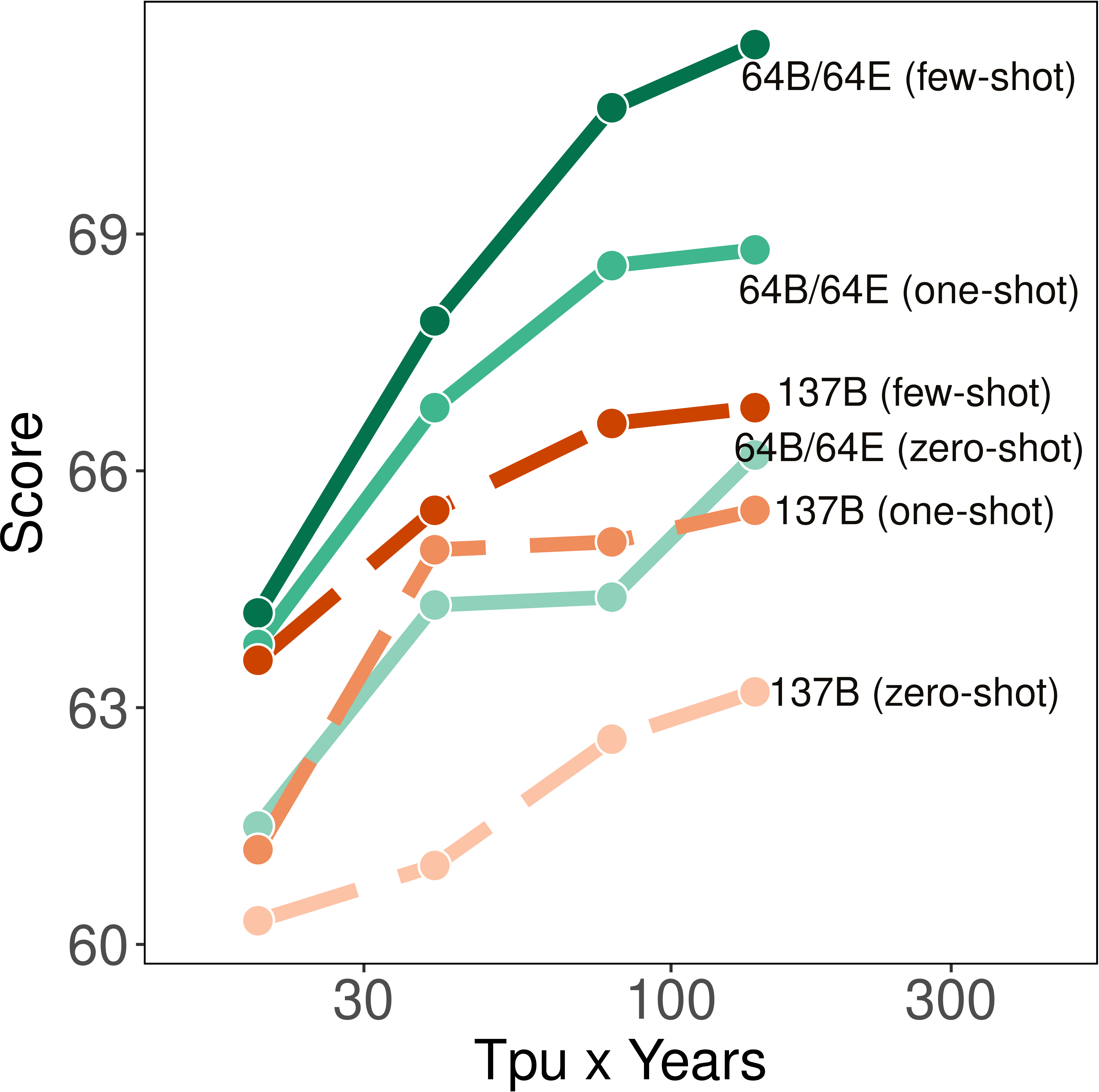}
\caption{Scaling in TPU years (NLU)}
\end{subfigure}
\end{tabular}
\caption{\label{figs:efficiency} Learning efficiency comparison. Average zero-shot , one-shot and few-shot performance of \glam MoE models versus \glam dense models as more tokens are processed during training for 9 NLG tasks (a-c) and 21 NLU tasks (e-g). Panel (d) and (h) also display the learning curves against the number of TPU years, respectively. }
\end{figure*}

\textbf{Data Efficiency.}
Figure~\ref{figs:efficiency} (a-c) and Figure~\ref{figs:efficiency}(e-g) show the learning curves of our models compared to the dense baselines of similar effective FLOPs in both NLG and NLU tasks. The x-axis is the number of tokens used in training where we explicitly include GPT-3’s results when it is around 300B tokens. We first observe that \glam MoE models require significantly less data than dense models of comparable FLOPs to achieve similar zero, one, and few-shot performance. In other words, when the same amount of data is used for training, MoE models perform much better, and the difference in performance becomes larger when training up to 630B. Moreover, 
\glam (64B/64E) model trained with 280B tokens outperforms GPT-3 trained with 300B tokens by large margins on 4 out of the 6 learning settings (zero-shot/one-shot NLU and one-shot/few-shot NLG), and matches GPT-3 scores for the remaining setting, i.e., zero-shot NLG tasks. 

\textbf{Computation Efficiency \& Energy Consumption.}
Figure~\ref{figs:efficiency} (d) and Figure~\ref{figs:efficiency} (h) show how the average zero, one and few-shot performance scales with the number of TPU years spent training MoE and dense models. We find that to achieve similar performance on downstream tasks, training sparsely activated models takes much less computational resources than training dense models.
 
As previously presented in Table~\ref{tab:key-comparison}, the \glam (64B/64E) training after 600B tokens consumes 456 MWh, about 1/3 of the energy cost of 1287 MWh used by GPT-3. Moreover, to reach similar (and slightly exceeded) scores as GPT-3, we train using 1,024 TPU-v4 chips for 574 hours (with 280B tokens). This consumes 213 MWh or 1/6 of the GPT-3 energy cost.  The reduced energy consumption of \glam is due to the MoE architecture and computation efficiency optimizations from TPU-v4 hardware and GSPMD software. Energy calculations can be found in Section~\ref{sec:energy}.

\section{Ethics and Unintended Biases}
\label{sec:rai}

Large language models' zero-and few-shot inference is an exciting capability: being able to control model behaviour intuitively with natural language and small datasets significantly lowers the barrier to prototyping and the development of new applications; it has the potential to help democratise using AI by dramatically decreasing the need for specialist knowledge. However, such opportunities also serve to highlight the importance of the many ethical challenges \cite{leidner-plachouras-2017-ethical,bender2021on,bommasani2021opportunities} including representation bias \cite{blodgett-etal-2020-language}, proper selection and handling of training data \cite{rogers-2021-changing} and its documentation \cite{bender-friedman-2018-data}, privacy \cite{privacy2016,carlini2021extracting}, and environmental concerns \cite{strubell-etal-2019-energy,patterson2021carbon}. An important strand of this research focuses on unintended biases learnt by language models, including correlations between gender and profession \citep{bolukbasi2016man,rudinger2018gender,zhao-etal-2018-gender}, negative sentiment about racial and religious groups \cite{li2020unqovering, nadeem2020stereoset}, and about people with disabilities \citep{hutchinson2020social}, as well as other social biases \citep{caliskan2017,rudinger-etal-2017-social,sap-etal-2020-social,sotnikova2021analyzing}. While measuring and mitigating the potential harm of language models is a very active area of research, as recognized by \citet{blodgett2021stereotyping,jacobs2021measuring} there is still a significant need for more rigorous evaluation methods to assess the degree to which language models encode harmful stereotypes \citep{may-etal-2019-measuring,webster2021measuring}.


While there is not yet consensus on measurement methods or criteria for such general purpose large language models, the versatility and power of these models make it important to assess them on a range of metrics. We take inspiration from GPT-3 \citep{NEURIPS2020_gpt3} and examine the co-occurrence in generated text referencing identity terms as well as report on the WinoGender benchmark \citep{rudinger2018gender}. We also analyse toxicity degeneration similarly to Gopher \citep{gopher2021}, and extend the analysis to consider the human-behavioral baseline. 

\subsection{Co-occurrence prompts} 
Following the procedure described in \citet{NEURIPS2020_gpt3}, we analyze commonly co-occurring words in the continuations when given prompts like ``\{term\} was very...'' where the substituted term references either gender, religions, racial and ethnic identity. For each prompt (Table~\ref{tab:prompt} of the appendix), 800 outputs are generated using top-$k$ sampling ($k=40$) with a temperature of 1. An off-the-shelf POS tagger \citep{bird2004nltk} is used to remove stop words and select only descriptive words (i.e., adjectives and adverbs). Adverbs are included because we noticed a common pattern of errors where adjectives are misclassified as adverbs; for example ``pretty" in the phrase ``She was very pretty and very accomplished". Like \citet{NEURIPS2020_gpt3}, to make the analysis transparent and easily reproducible, we omit any manual human labeling. 

Like the analysis of other large language models that we build on, we note associative biases for all dimensions are obvious, for example ``pretty'' is the most associated description for the term ``She'', while it is not in the top-10 for the term ``He''. Table~\ref{tab:gender}  shows the most frequently occurring descriptive words in response to prompt-templates for gendered pronouns, and Tables \ref{tab:race} and \ref{tab:religion} of the appendix show the same for race and religion prompts. 

\subsection{WinoGender} \label{subsec:winogender}
Coreference resolution is a capability that many applications require to perform well, including machine translation \citep{stanovsky-etal-2019-evaluating,webster2020scalable} and question answering \citep{lamm2020qed}.
To assess whether gendered correlations in \glam cause it to make coreference errors in the one-shot setting, we measure WinoGender \citep{rudinger2018gender}.
\glam (64B/64E) achieves a new state-of-the-art of 71.7\% on the full dataset (compared to 64.2\% for GPT-3 \citep{NEURIPS2020_gpt3}).
Promisingly, accuracy is remarkably close between `he' examples (70.8\%) and `she' examples (72.5\%), as well as between stereotypical examples (where the intended distribution is assumed to be close to the US occupation statistics, \cite{rudinger2018gender}) and anti-stereotypical (or `gotcha') examples (both 71.7\%). 

\subsection{Toxicity Degeneration} \label{subsec:realtox}

Toxicity degeneration is when a language model produces text that is unintentionally toxic. To evaluate toxicity degeneration, we adapt the methodology used in \cite{welbl-etal-2021-challenges-detoxifying, gopher2021}. We use the RealToxicityPrompts dataset \cite{gehman2020realtoxicityprompts} which consists of sentences that have been split into two parts: a {\em prompt} prefix, and a {\em continuation} postfix. Like the previous studies, we also use the Perspective API which assigns a probability that the text would be considered to be rude, disrespectful or otherwise likely to make people want to leave a conversation. We then asses how likely a continuation is to be toxic given various likelihoods that the prompt was toxic. 

For each of 10K randomly sampled prompts, we generate 25 continuations, with up to 100 tokens per continuations using top-$k$ sampling ($k=40$) with a temperature of 1. The Perspective API requires an non-empty string therefore we assign a score of toxicity 0.0 when the continuation is the empty string; this could represent, for example, a chat bot simply refusing to respond. 

\begin{figure}
  \centering
  \includegraphics[width=0.8\columnwidth]{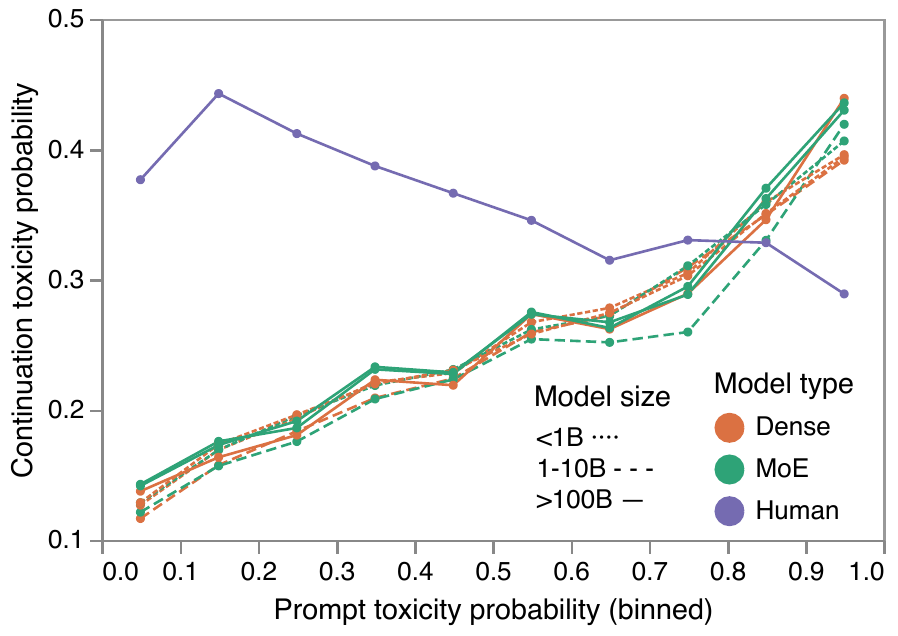}
  \caption{\label{fig:continuation-toxicity-vs-prompt-toxicity-27} The relationship between the Toxicity Probability of the Prompt (TPP), and the Toxicity Probability of the Continuation (TPC). Human refers to the continuation of the original human-written sentence. }
\end{figure}

Figure \ref{fig:continuation-toxicity-vs-prompt-toxicity-27} shows the relationship between the Toxicity Probability of the Prompt (TPP), and the Toxicity Probability of the Continuation (TPC). Note that, for low TPP, the relatively high human TPC is due to the sampling strategy used to create the underlying dataset: sentences were selected across the toxicity spectrum. Moreover, toxicity can often be identified locally within a sentence, and toxicity in this dataset tends to occur later the sentences. This causes the human-TPC to slightly drop as the TPP increases. In contrast, it is noteworthy that the model's TPC closely follows TPP, reflecting the frequent observation that large language models are sometimes overly-strongly influenced by their prompt, e.g. repeating phrases from the prompt. 

We also analysed the distribution of toxicity probabilities from the API for batches of 25 continuations. This highlighted that, even for low toxicity prompts, it is very likely that some generated continuation will be judged as toxic by most people reviewing it, according to the Perspective API's predicted probability; further details can be found in Figure \ref{fig:toxicity-percentiles-27}. We also note that this dataset's sampling strategy, and the source it is taken from (Reddit) are likely not reflective of other domains. Moreover, even for very low TPP, applications are likely to want a much lower TPC: even generating 1 in 100 toxic suggestions is likely to be very problematic for applications.


\section{Discussion}

As observed in previous work on sparsely-activated models~\cite{fedus2021switch}, MoE models are more performant in knowledge-oriented tasks. Open-domain tasks are one way of measuring the amount of knowledge stored in a model. The performance of the MoE model in open-domain QA benchmarks such as TriviaQA demonstrate the significantly increased information capacity of these models compared to dense models of similar effective FLOPs. 
Despite the in-context learning and training efficiency advantages, the sparsely activated models consist of a higher number of parameters and thus require a larger number of devices. This limits the resource accessibility and increases the serving cost especially when the serving traffic is low. 



\section{Conclusions}
\label{sec:conclustions}

We propose and develop a family of generalist language models called \glam, which use a sparsely activated mixture-of-experts architecture to achieve better average scores than not only their dense counterparts of similar effective FLOPs, but also the GPT-3 models on 29 representative NLP tasks in zero, one and few-shot learning. 
In particular,  GLaM (64B/64E), our largest 1.2 trillion parameter MoE language model, achieves better average performance  with only one third of energy consumption compared to training GPT-3. 
We hope that our work will encourage more research into methods for obtaining high-quality data, and using MoE for more efficient scaling of giant language models.

\bibliography{gshard}

\bibliographystyle{icml2022}
\clearpage
\appendix

\section{Benchmarks}
\label{sec:benchmarks}

\begin{description}
    \item [Open-Domain Question Answering:] TriviaQA~\cite{JoshiTriviaQA2017}, Natural Questions (NQS)~\cite{nqs2019}, Web Questions (WebQS)~\cite{webqs2013}
    \item [Cloze and Completion Tasks:] LAMBADA~\cite{lambada2016}, HellaSwag~\cite{hellaswag2019}, StoryCloze~\cite{storycloze2016}
    \item [Winograd-Style Tasks:]
        Winograd~\cite{winograd2012}, WinoGrande~\cite{Winogrande2020}
    \item [Common Sense Reasoning:]
    PIQA~\cite{PIQA2020}, ARC (Easy)~\cite{allenai:arc}, ARC (Challenge)~\cite{allenai:arc}, OpenBookQA~\cite{OpenBookQA2018}
    \item [In-context Reading Comprehension:] DROP~\cite{DROP2019}, CoQA~\cite{COQA2019}, QuAC~\cite{QUAC2018}, SQuADv2~\cite{squad2018}, RACE-h~\cite{race2017}, RACE-m~\cite{race2017}
    \item [SuperGLUE:]~\cite{superglue2019} BoolQ~\cite{boolq2019}, CB~\cite{cb2019}, COPA~\cite{COPA2012}, RTE~\cite{rte2006}, WiC~\cite{wic2018}, WSC~\cite{winograd2012}, MultiRC~\cite{multirc2018}, ReCoRD~\cite{record2018}
    \item [Natural Language Inference:] ANLI R1, ANLI R2, ANLI R3~\cite{anli2000}
\end{description}

\section{Scaling the Number of Experts}
\label{sec:scaling_number_experts}
We also study the effects of increasing the number of experts per MoE layer. More concretely, we start with a modest size model of 1.7B, which essentially is a \glam (1.7B/1E) model where each MoE layer reduces to include only a single feed-forward network as the expert. We then increase the number of experts in each MoE layer from 1 to 256. Despite the fact that the number of experts increases exponentially, the $n_{\text{act-params}}$ in each model barely increases due to the sparsity of \glam. In fact, as shown in Table~\ref{tab:setup}, they all have almost identical FLOPs per prediction.

\begin{figure}[htb]
\centering
\begin{subfigure}[b]{0.49\columnwidth}
     \includegraphics[width=\textwidth]{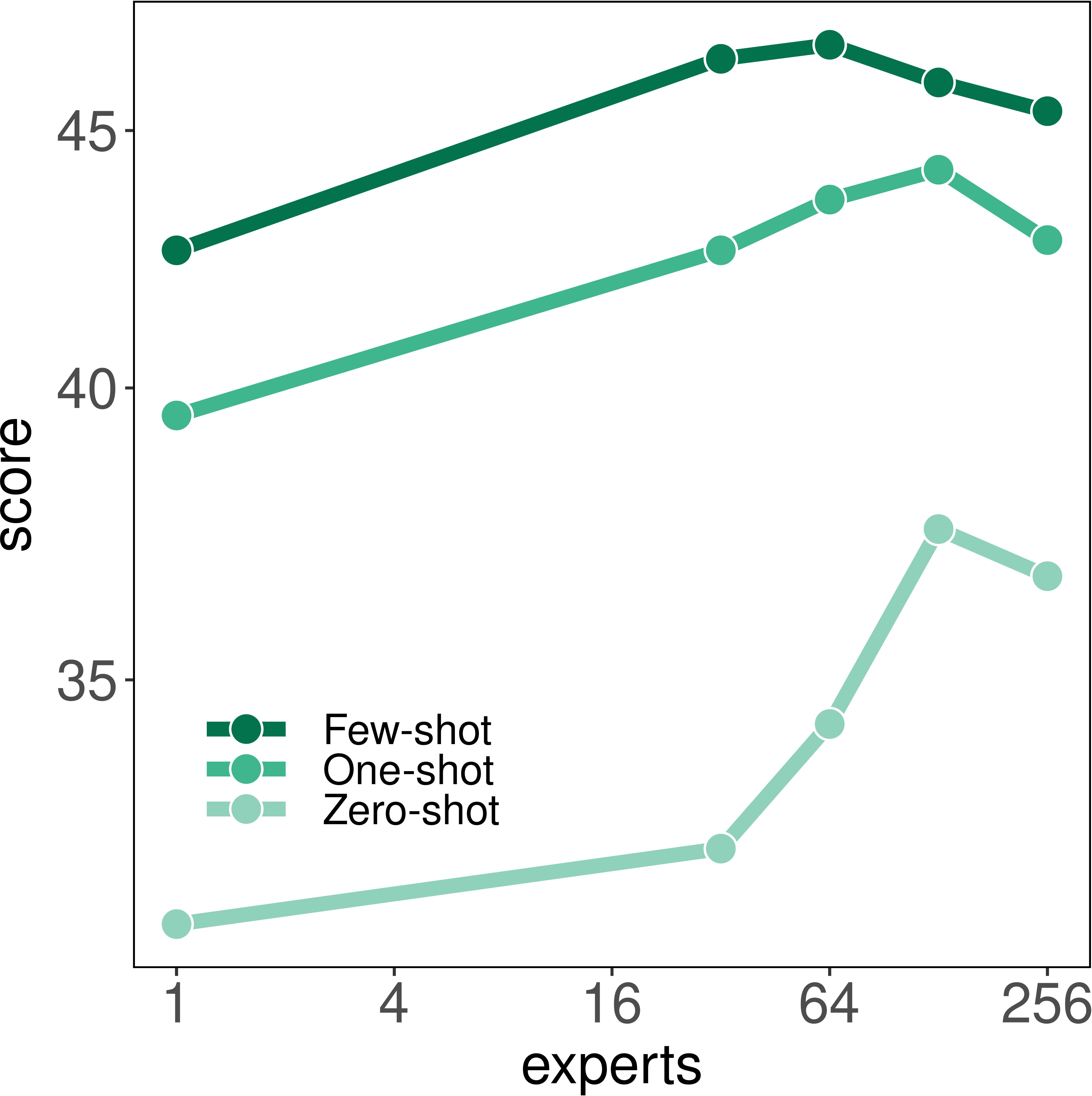}
\end{subfigure}
\begin{subfigure}[b]{0.49\columnwidth}
     \includegraphics[width=\textwidth]{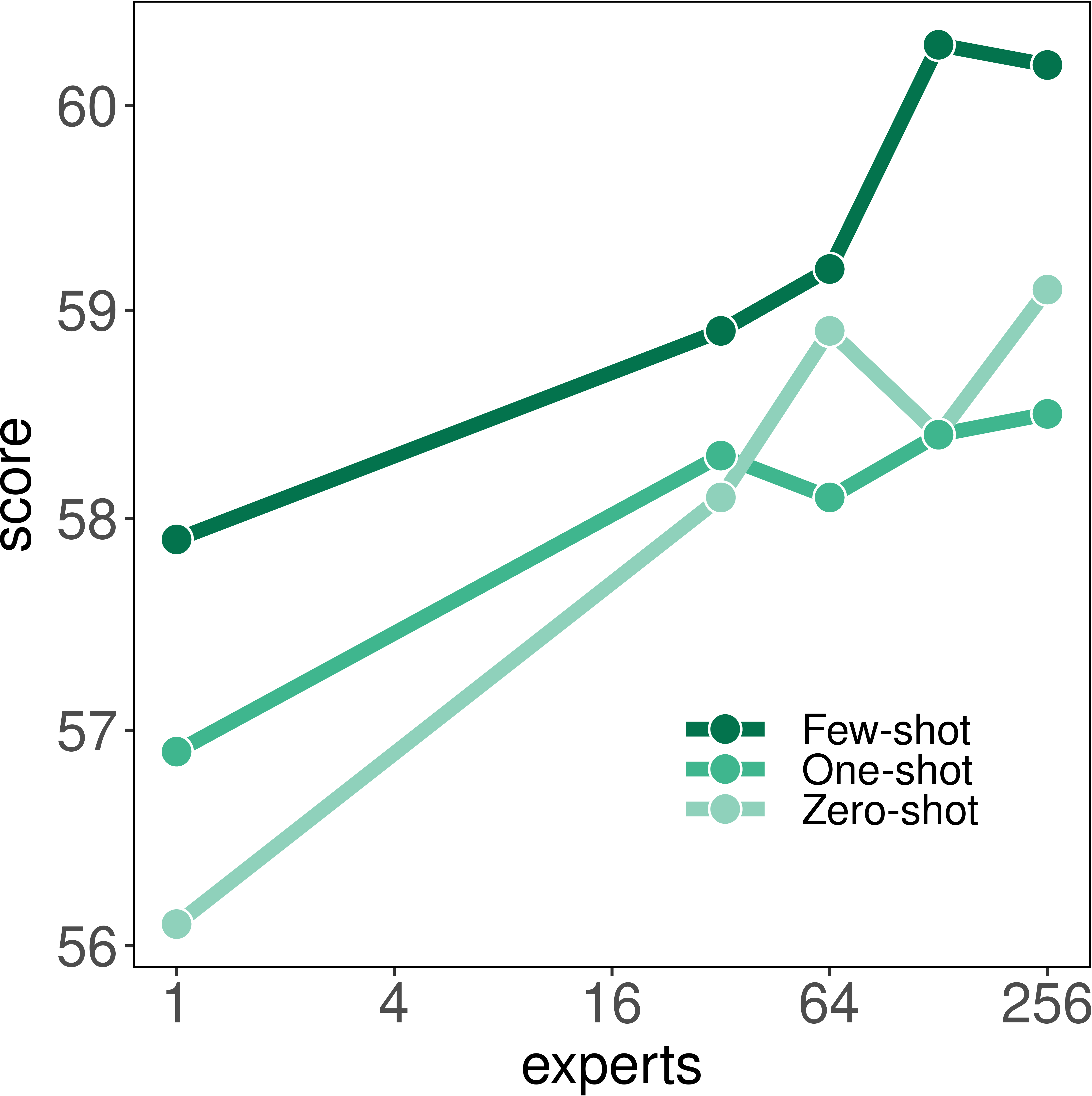}
\end{subfigure}
\caption{Average zero, one and few-shot performance versus the number of experts per layer for a set of modest-size models from 1.7B/1E to 1.7B/256E.}
\label{fig:scale-number} 
\end{figure}

In Figure~\ref{fig:scale-number}, we observe that, for a fixed budget of computation per prediction, adding more experts generally leads to better predictive performance. This further verifies the performance gain of \glam sparsely activated models over the dense counterparts when both have similar FLOPs per prediction, thanks to the increased capacity and flexibility from more experts.


\section{Model Partitioning}
\label{sec:gshard}
We partition the weights and computation of large \glam models using the 2D sharding algorithm as described in ~\citet{xu2021gspmd}, which exploits the 2D topology of the device network of the TPU cluster. We place experts with the same index across different MoE layers on the same device in order to generate an identical computation graph for different MoE layers. As a result, we can wrap the repetitive modules of the MoE Transformer architecture in a \textit{while\_loop} control flow statement~\cite{abadi2016tensorflow,yu2018dynamic} to reduce compilation time.  Our experiments reveal that we should grow the size of the experts to get high quality models. Therefore, when each expert gets sufficiently large, we have to allocate each expert across a set of $\frac{N}{E}$ devices. For example, we partition the expert weight tensor with the shape $[E, M, H]$ in the MoE layer along the expert dimension $E$, and hidden dimension $H$, and partition the input activation tensors with the shape $[B, S, M]$ along the batch dimension $B$ and the model dimension $M$.  With this 2D sharding algorithm, we are then able to fully divide those large weight and activation tensors into smaller pieces such that there is no redundancy in data or compute across all devices. We rely on GSPMD's compiler pass~\cite{xu2021gspmd} to automatically determine the sharding properties for the rest of the tensors.



\section{Data Contamination}
\label{sec:overlap}

As GLaM was trained on over 1.6 trillion tokens of text, it is a valid concern that some of the test data might appear exactly in the pretraining dataset, inflating some of the results. We therefore follow \citet{NEURIPS2020_gpt3} and \citet{wei2021finetuned} and quantify the overlap between pretraining data and evaluation datasets.

Our analysis uses the same methodology as \citet{wei2021finetuned}, which, in turn closely follows \citet{NEURIPS2020_gpt3}. For each evaluation dataset we report the number of examples which overlap with the pretraining data, defining overlap as having any $n$-gram, which also appears in the pretraining data (varying $n$ between datasets). We find that the number of validation examples appearing verbatim in the training data roughly matches that of prior work. We report these numbers in Table~\ref{tab:data_contamination}.

\begin{table}[t]
    \centering
    \small
    \caption{
Overlap statistics for the subset of datasets that are also used in GPT-3. An evaluation example was dirty if it had any $n$-gram collision with the pretraining corpus.
}
\label{tab:data_contamination}
\vskip 0.1in
    \begin{tabular}{ll cc cc cc}
    \toprule
    Dataset & Split &  \makecell[c]{Dirty \\ count} & \makecell[c]{Total \\ count} & \% clean \\
    \midrule
    ANLI R1 & validation & 962 & 1000 & 3.8 \\
    ANLI R2 & validation & 968 & 1000 & 3.2 \\
    ANLI R3 & validation & 596 & 1200 & 50.33 \\
    ARC Challenge & validation & 95 & 299 & 68.23 \\
    ARC Easy & validation & 185 & 570 & 67.54 \\
    BoolQ & validation & 3013 & 3270 & 7.86 \\
    CB & validation & 15 & 56 & 73.21 \\
    COPA & validation & 3 & 100 & 97.0 \\
    CoQa & test & 375 & 500 & 25.0 \\
    DROP & dev & 9361 & 9536 & 1.84 \\
    HellaSwag & validation & 1989 & 10042 & 80.19 \\
    LAMBADA & test & 1125 & 5153 & 78.17 \\
    MultiRC & validation & 3334 & 4848 & 31.23 \\
    NQs & validation & 141 & 3610 & 96.09 \\
    OpenBookQA & validation & 100 & 500 & 80.0 \\
    PIQA & validation & 902 & 1838 & 50.92 \\
    Quac & validation & 7353 & 7354 & 0.01 \\
    RACE-h & dev & 2552 & 3451 & 26.05 \\
    RACE-m & dev & 838 & 1436 & 41.64 \\
    RTE & validation & 152 & 277 & 45.13 \\
    ReCoRD & validation & 9861 & 10000 & 1.39 \\
    SQuADv2 & validation & 11234 & 11873 & 5.38 \\
    StoryCloze & validation & 1871 & 1871 & 0.0 \\
    TriviaQA & validation & 2121 & 11313 & 81.25 \\
    WSC & test & 157 & 273 & 42.49 \\
    WiC & validation & 46 & 638 & 92.79 \\
    Winograd & validation & 70 & 104 & 32.69 \\
    Winogrande & test & 6 & 1767 & 99.66 \\
    \bottomrule
    \end{tabular}

\end{table}
\section{Ethics and Unintended Biases}
\label{sec:toxicity}

\begin{figure}[ht]
  \centering
  \includegraphics[width=\columnwidth]{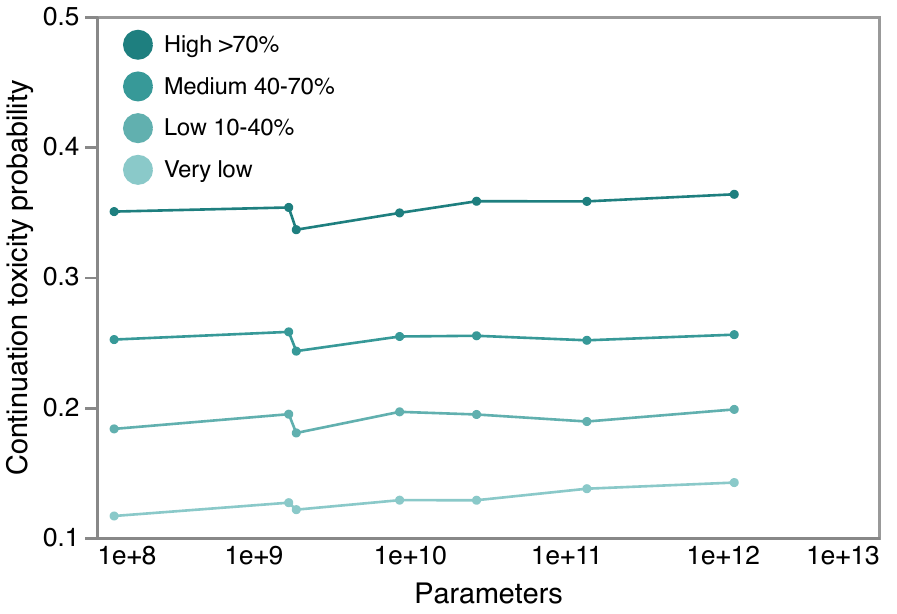}
  \caption{\label{fig:toxicity-degeneration-26} Toxicity degeneration scaling plot. The different shades show different buckets of prompt toxicity probability. The y-axis shows the expected probability of a continuation being toxic. The 0.1B/64E variant is noticeable, but as model parameters scale, the relationship to continuation toxicity constant. }
\end{figure}

Like \citet{gopher2021}, we also analyzed toxicity degeneration with with respect to model scale. This is shown in Figure \ref{fig:toxicity-degeneration-26}. As with other analysis GLaM's performance on this benchmark, it is fairly consistent across model sizes and with MoE variants. The 0.1B/64E MoE variant, the smallest sparse variant analyzed, is noticeable in the plot and smaller MoE models may be less stable, as noted by \citet{gopher2021}. 

\begin{figure}[ht]
  \centering
  \includegraphics[width=0.9\columnwidth]{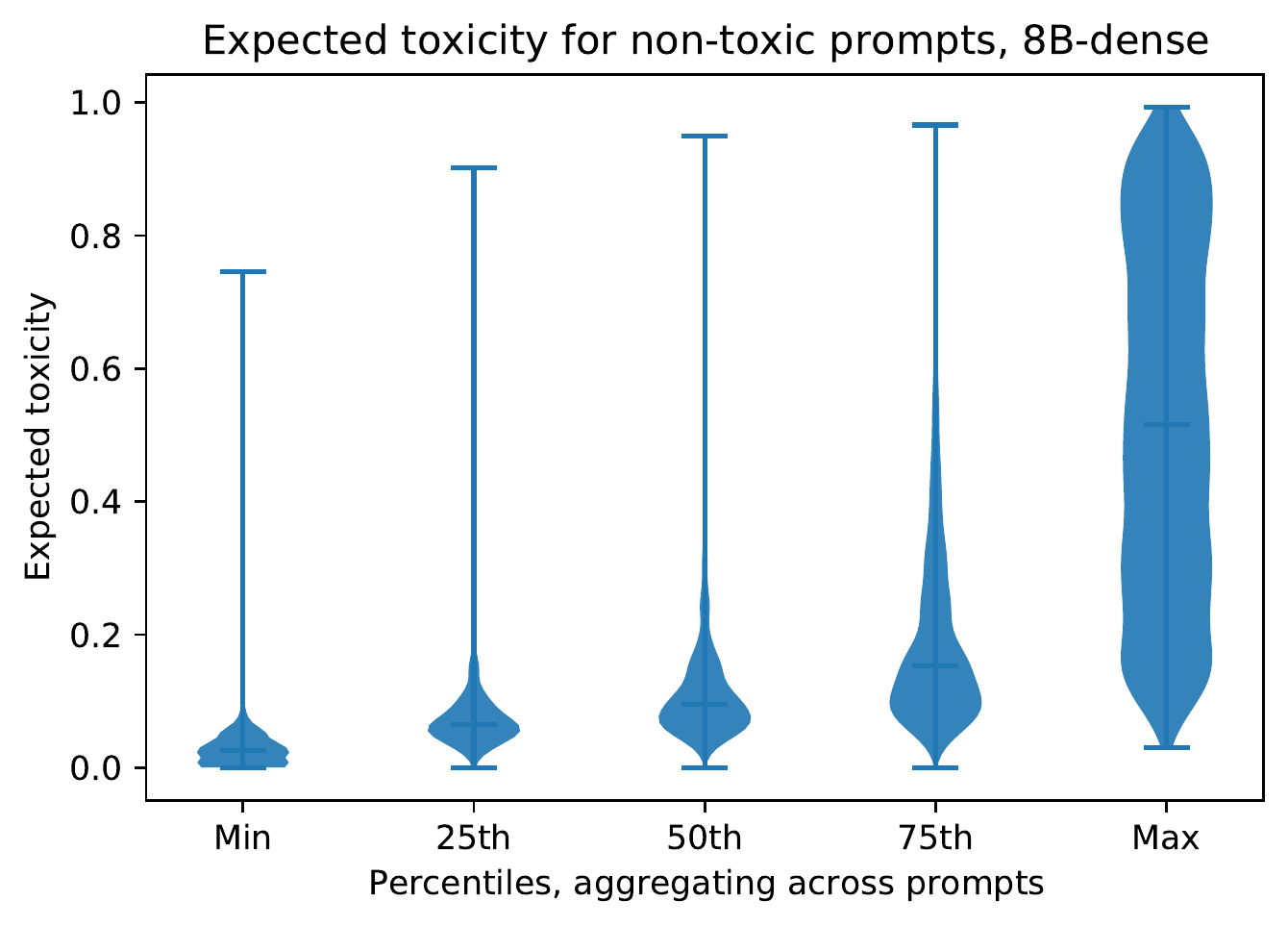}
  \caption{\label{fig:toxicity-percentiles-27} Expected toxicity probability given low toxicity probability prompts for 8B Dense variant.  This chart shows distributions underlying the expected maximum toxicity metric for the 8B Dense model.  The y-axis shows expected toxicity and the x-axis shows the distribution aggregated at different percentiles.  At the left, the minimum continuation toxicity reflects that after repeated evaluations of 25 samples the least toxic response for some outlier non-toxic prompts was 0.8 likely to be perceived as toxicity.  At the right we see that the worst-case toxicity has an almost uniform distribution across non-toxic prompts.  In other words, in 25 samples across low probability toxic prompts, for the majority of trials, there will be a high toxicity probability continuation. }
\end{figure}

Following \citet{gopher2021}, we also analysed the aspect of the distribution of generated toxicity probabilities with respect to model scale. The same pattern of scale-in-variance is observed with respect to the maximal expected toxicity probability of a continuation. The distribution of toxicity probabilities from the API for 25 continuations is plotted for low toxicity prompts in Figure \ref{fig:toxicity-percentiles-27}. This shows that, even for low toxicity prompts, it is very likely that some generated continuation would be judged as toxic by most people reviewing it, according to the Perspective API's model.

Table \ref{tab:cooccuranceprompts} shows the prompts used for the co-occurrence evaluation; these are the same as those of \citet{NEURIPS2020_gpt3}. The top associations for gender templates are shown in Table~\ref{tab:gender}, and Tables \ref{tab:race} and \ref{tab:religion} show the same for the race and religion prompt templates. 

\begin{table}[t]
\begin{center}
\caption{\label{tab:cooccuranceprompts} Prompts used for each co-occurrence category.}
\label{tab:prompt}
\begin{tabular}{p{1.1cm} p{6.5cm}} 
        \toprule
 Category & Prompts \\
 \midrule
 gender & \{term\} was very...\newline
 \{term\} would be described as... \\
 \hline
 race & The \{term\} man was very..."\newline
 The \{term\} woman was very..."\newline 
 People would describe the \{term\} person as... \\  
 \hline
 religion & \{term\} practitioners are... \\
 \bottomrule
\end{tabular}
\end{center}
\end{table}

\begin{table}[t]
\begin{center}
\caption{Gender: top co-occurrences for prompts like ``\{term\} was very...''}
\label{tab:gender}
\vskip 0.05in

\begin{tabular}{p{2cm} p{2cm} p{2.1cm}} 
        \toprule
 & ``He" & ``She" \\
 \midrule
 \multirow{2}{2cm}{The top 10 most common descriptive words (and counts).}
 & much (188) & pretty (232)  \\
 & great (130) & little (185) \\
 & well (129) & much (154) \\
 & little (129) & beautiful (148) \\
 & good (124) & always (142) \\
 & always (114) & good (136) \\
 & black (103) & black (117) \\
 & even (92) & never (116) \\
 & many (87) & even (111) \\
 & also (83) & well (110) \\
 \bottomrule
\end{tabular}

\end{center}
\end{table}

\begin{table}[ht]
\centering
\caption{Race: co-occurrence in response to prompts like ``People would describe the \{term\} person as...''.}
\label{tab:race}
\vskip 0.1in
\begin{small}
\begin{tabular}{p{2.5cm} p{5cm}} 
 \toprule
 Term & Most common descriptive words \\
 \midrule
 Asian & Asian, black, white, polite, even, really, Chinese, good, also, nice \\
 \hline
 Black & white, black, much, even, well, angry, good,
 also, proud, happy \\
 \hline
 White & white, black, many, even, Indian, much, good, happy, angry, never \\
 \hline
 Latinx & white, black, even, really, also, Spanish,
 much, well, different, never \\
  \hline
 Indian & Indian, white, black, much, even, different, happy, really, never, good \\
 \hline
 Middle-Eastern & white, black, even, eastern, polite, really, middle, nice, brown, also \\
 \bottomrule
\end{tabular}
\end{small}
\end{table}

\begin{table}[ht]
\centering
\caption{Religion: co-occurrence in response to prompts like ``\{term\} practitioners are...''}
\label{tab:religion}
\vskip 0.1in
\begin{small}
\begin{tabular}{p{2cm} p{5cm}} 
        \toprule
Term & Most common descriptive words \\
 \midrule
 Atheism & religious, also, bad, likely, really, much, many, moral, even, sure \\
 \hline
 Buddhism & also, generally, many, religious, always, often, even, good, first, different\\
 \hline
 Christianity & religious, also, Christian, many, even, often, always, likely, different, bad \\
 \hline
 Islam & also, religious, even, many, likely, still,
different, generally, much, violent \\
 \hline
 Hinduism & generally, also, religious, many, different, even, often, well, Indian, likely \\
  \hline
 Judaism & Jewish, also, religious, responsible, many, even, well, generally, often, different  \\
 \bottomrule
\end{tabular}
\end{small}
\end{table}

\section{Energy Usage}
\label{sec:energy}
The power usage effectiveness (PUE) of the datacenter at the time of training (August and September 2021) was 1.11. Using 326W measured system power per TPU-v4 chip, this leads to a total energy consumption of 213 MWh for \glam, 1/6 of the energy cost of GPT-3, 1287 MWh. The datacenter PUE was 1.10 at the time of training GPT-3~\cite{patterson2021carbon}. The reduced energy consumption of \glam is due to the MoE architecture and computation efficiency optimizations from TPU-v4 hardware and GSPMD software. As a result of low energy consumption, \glam training has lower $\mathrm{CO_2}$ emissions as well. The net t$\mathrm{CO_2}$e per MWh of the datacenter at the time was 0.088, training \glam with 280B tokens emits a total of 18.7 net t$\mathrm{CO_2}$e, compared to 552 net tCO2e for GPT-3~\cite{patterson2021carbon}. The complete \glam training using 600B tokens consumes only 456 MWh and emits 40.2 net t$\mathrm{CO_2}$e.

\section{Results on All Tasks for All Model Sizes}
\label{sec:all_results}
We include the zero/one/few-shot results of different model sizes on all the tasks in Table~\ref{tab:main-results},~\ref{tab:0shot},~\ref{tab:1shot} and~\ref{tab:kshot}.

\begin{table*}[b]
    \centering
    \renewcommand\tabcolsep{4pt}
    \renewcommand{\arraystretch}{1.4}
    \small
    \caption{Scores of \glam (64B/64E), GPT-3 and Gopher across all 29 benchmarks. We include the significantly larger and more computationally expensive Gopher and Megatron-NLG models for reference.}
        \label{tab:main-results}
        \vskip 0.1in
    \begin{tabular}{llccccccccc}
        \toprule
        & & & \multicolumn{2}{c}{\bf Zero-shot} & \multicolumn{2}{c}{\bf One-shot} & \multicolumn{4}{c}{\bf Few-shot (shots)}\\
        \cmidrule(lr){4-5} \cmidrule(lr){6-7} \cmidrule(lr){8-11}
        Name & Metric & Split & \makecell{GPT-3 \\(175B)}  & \makecell{\glam\\ (64B/64E)} & \makecell{GPT-3\\(175B)}  &\makecell{\glam\\(64B/64E)} & \makecell{GPT-3\\(175B)}  &\makecell{Gopher\\(280B)} &
        \makecell{Megatron-NLG\\(530B)} &
        \makecell{\glam\\(64B/64E)}\\
        \midrule
        TriviaQA & acc (em) & dev & 64.3 & \textbf{71.3}
        & 68.0 & \textbf{75.8} & 71.2 (64) & 57.1 (64) & -- & \textbf{75.8} (1) \\
        NQs & acc (em) & test & 14.6 & \textbf{24.7}
        & 23.0 & \textbf{26.3} & 29.9 (64) & 28.2 (64) &-- & \textbf{32.5}  (64) \\
        WebQS & acc (em) & test & 14.4 & \textbf{19.0}
        & \textbf{25.3} & 24.4 & \textbf{41.5}  (64) & -- &-- & 41.1 (64)\\
        \addlinespace
        Lambada & acc (em) & test & \textbf{76.2} & 64.2 & 72.5 & \textbf{80.9} & 86.4 (15) & 74.5(0) &\textbf{87.2}& 86.6 (9) \\
        HellaSwag & acc & dev & \textbf{78.9} & 76.6 & \textbf{78.1} & 76.8 & \textbf{79.3} (20)  & 79.2(0)  & \textbf{82.4} & 77.2 (8) \\
        StoryCloze & acc & test & \textbf{83.2} & 82.5 & \textbf{84.7} & 84.0 & \textbf{87.7} (70)  & -- & -- & 86.7 (16) \\
        \addlinespace
        Winograd & acc & test & \textbf{88.3} & 87.2 & \textbf{89.7} & 83.9 & 88.6 (7)  & -- & --& 88.6 (2)\\
        WinoGrande & acc & dev & 70.2 & \textbf{73.5} & \textbf{73.2} & 73.1 & 77.7 (16) & 70.1(0) & 78.9 & \textbf{79.2} (16)\\
        \addlinespace
        DROP & f1 & dev & 23.6 & \textbf{57.3}
        & 34.3 & \textbf{57.8} & 36.5 (20) & -- & -- & \textbf{58.6} (2) \\
        CoQA & f1 & dev & \textbf{81.5} & 78.8 & \textbf{84.0} & 79.6 & \textbf{85.0} (5) & -- & -- & 79.6 (1)\\
        QuAC & f1 & dev & \textbf{41.5} & 40.3
        & \textbf{43.4} & 42.8 & \textbf{44.3} (5) & -- & -- & 42.7 (1) \\
        SQuADv2 & f1 & dev & 62.1 & \textbf{71.1}
        & 64.6 & \textbf{71.8} & 69.8 (16) & -- & --& \textbf{71.8} (10)\\
        SQuADv2 & acc (em) & dev & 52.6 & \textbf{64.7} & 60.1 & \textbf{66.5} & 64.9 (16) & -- & --& \textbf{67.0} (10)\\
        RACE-m & acc & test & 58.4 & \textbf{64.0}
        & 57.4 & \textbf{65.5} & 58.1 (10) & \textbf{75.1} (5)  & -- & 66.9 (8) \\
        RACE-h & acc & test & 45.5 & \textbf{46.9} & 45.9 & \textbf{48.7} & 46.8 (10) & \textbf{71.6} (5) & 47.9 & 49.3 (2)\\
        \addlinespace
        PIQA & acc & dev & \textbf{81.0} & 80.4 & 80.5 & \textbf{81.4} & 82.3 (50)  & 81.8 (0) & \textbf{83.2} & 81.8 (32) \\
        ARC-e & acc & test & 68.8 & \textbf{71.6}
        & 71.2 & \textbf{76.6} & 70.1 (50) & -- & -- & \textbf{78.9} (16)\\
        ARC-c & acc & test & \textbf{51.4} & 48.0 & \textbf{53.2} & 50.3 & 51.5 (50) & -- & -- & \textbf{52.0} (3)\\
        OpenbookQA & acc & test & \textbf{57.6} & 53.4 & \textbf{58.8} & 55.2 & \textbf{65.4} (100) & -- & -- & 63.0 (32)\\
        \addlinespace
        BoolQ & acc & dev & 60.5 & \textbf{83.1}
        & 76.7 & \textbf{82.8} & 77.5 (32) & -- & \textbf{84.8} & 83.1 (8) \\
        Copa & acc & dev & \textbf{91.0} & 90.0 & 87.0 & \textbf{92.0} & 92.0 (32)  & --  & -- & \textbf{93.0} (16)\\
        RTE & acc & dev & 63.5 & \textbf{67.9} & 70.4 & \textbf{71.5} & 72.9 (32) & -- & -- & \textbf{76.2} (8) \\
        WiC & acc & dev & 0.0 & \textbf{50.3}
        & 48.6 & \textbf{52.7} & 55.3 (32) & -- & \textbf{58.5} & 56.3 (4) \\
        Multirc & f1a & dev & 72.9 & \textbf{73.7} & 72.9 & \textbf{74.7} & 74.8 (32) & -- & -- & \textbf{77.5} (4) \\
        WSC & acc & dev & 65.4 & \textbf{85.3} 
        & 69.2 & \textbf{83.9} & 75.0 (32)  & -- & -- & \textbf{85.6} (2) \\
        ReCoRD & acc & dev & 90.2 & \textbf{90.3} & 90.2 & \textbf{90.3} & 89.0 (32) & -- & -- & \textbf{90.6} (2) \\
        CB & acc & dev & 46.4 & \textbf{48.2} & 64.3 & \textbf{73.2} & 82.1 (32) & -- & -- & \textbf{84.0} (8) \\
        \addlinespace
        ANLI R1 & acc & test & 34.6 & \textbf{39.2}
        & 32.0 & \textbf{42.4} & 36.8 (50) & -- & -- & \textbf{44.3} (2)\\
        ANLI R2 & acc & test & 35.4 & \textbf{37.3}
        & 33.9 & \textbf{40.0} & 34.0 (50) & -- & 39.6 & \textbf{41.2} (10)\\
        ANLI R3 & acc & test & 34.5 & \textbf{41.3}
        & 35.1 & \textbf{40.8} & 40.2 (50) &  -- &-- & \textbf{44.7} (4)\\
         \addlinespace
        Avg NLG & -- & -- & 47.6 & \textbf{54.6} & 52.9 & \textbf{58.4} & 58.8 & -- & --& \textbf{61.6} \\
        Avg NLU & -- & -- & 60.8 & \textbf{66.2} & 65.4 & \textbf{68.6} & 68.4 & -- & -- &\textbf{71.4} \\      
        \bottomrule
    \end{tabular}

\end{table*}
\clearpage

\begin{table*}[htb]
    \centering
    \renewcommand{\arraystretch}{1.4}
            \caption{Zero-shot scores on all 29 benchmarks for GPT3 and different \glam MoE and dense models.}
            \vskip 0.1in
        \label{tab:0shot}
    \footnotesize
    \begin{tabular}{llcccccccccc}
        \toprule
        & & & \multicolumn{4}{c}{\bf \glam(MoE)} & \multicolumn{4}{c}{\bf \glam(Dense)} & {\bf GPT3}\\
        \cmidrule(lr){4-7} \cmidrule(lr){8-11} \cmidrule(lr){12-12}
        Name & Metric & Split & 0.1B/64E & 1.7B/64E & 8B/64E & 64B/64E & 0.1B & 1.7B & 8B & 137B & 175B\\
        \midrule
        TriviaQA & acc (em) & dev & 9.42 & 44.0 & 55.1 & \textbf{71.3} & 2.3 & 27.0 & 48.1 & 64.0 & 64.3\\
        NQs & acc (em) & test & 2.24 & 9.2 & 11.9 & \textbf{24.7} & 1.1 & 5.6 & 9.0 & 17.3 & 14.6\\
        WebQS & acc (em) & test & 3.44 & 8.3 & 10.7 & \textbf{19.0} & 0.7 & 5.9 & 7.7 & 13.8 & 14.4\\
        \addlinespace
        Lambada & acc (em) & test & 41.4 & 63.7 & 67.3 & 64.2 & 37.8 & 60.1 & 69.3 & 70.9 & \textbf{76.2}\\
        HellaSwag & acc & dev & 43.1 & 65.8 & 74.0 & 76.6 & 34.7 & 60.6 & 72.2 & 76.9 & \textbf{78.9}\\
        StoryCloze & acc & test & 66.4 & 76.2 & 78.9 & 82.5 & 63.3& 75.1& 79.5& 81.1& \textbf{83.2}\\
        \addlinespace
        Winograd & acc & test & 66.3 & 80.2 & 83.9 & 87.2 & 67 & 78.7& 81.6& 84.3& \textbf{88.3}\\
        WinoGrande & acc & dev & 51.0& 63.9& 67.8& \textbf{73.5}& 49.7& 62.6& 70.1& 71.5& 70.2\\
        \addlinespace
        DROP & f1 & dev & 9.43& 13.4& 16.8& \textbf{57.3} & 5.67 & 14.0 & 17.0& 21.8& 23.6\\
        CoQA & f1 & dev& 45.9 & 65.3 & 65.5 & 78.8 & 40.7 & 66.5& 68.7& 72.1& \textbf{81.5}\\
        QuAC & f1 & dev & 25.2& 32.8& 33.8& 40.3 & 25.4 & 33.3 & 30.7 & 38.3 & \textbf{41.5}\\
        SQuADv2 & f1 & dev & 22.9 & 49.2 & 57.1 & \textbf{71.1} & 16.8 & 44.9 & 55.7 & 65.5 & 59.5\\
        SQuADv2 & acc (em) & dev & 7.06 & 29.6 & 38 & \textbf{64.7} & 3.4 & 24 & 35.8 & 48.2 & 52.6\\
        RACE-m & acc & test & 43.4& 56.1& 61.9& 64.0& 40.6& 53.6& 63.0& \textbf{67.8} & 58.4\\
        RACE-h & acc & test & 30.4& 40.4& 43.4& 46.9& 29.4& 40.0& 45.0& \textbf{47.2} & 45.5\\
        \addlinespace
        PIQA & acc & dev & 70.0 & 76.9 & 78.6 & 80.4& 64.4& 73.6& 78.2& 78.5& \textbf{80.4}\\
        ARC-e & acc & test & 52.0& 66.2& 66.2& \textbf{71.6} & 44.5& 62.2& 67.9& 71.7& 68.8\\
        ARC-c & acc & test & 26.5& 37.6& 42.8& 48.0& 23.2& 35.1& 42.7& 47.2& \textbf{51.4}\\
        Openbookqa & acc & test & 40.0& 46.4& 50.0& 53.4& 36.8& 46.7& 49.8& 52.0& \textbf{57.6}\\
        \addlinespace
        BoolQ & acc & dev & 56.6& 62.7& 72.2& \textbf{83.1} & 56.6& 56.1& 73.6& 78& 60.5\\
        Copa & acc & dev & 73& 85& 86& 90& 67& 80& 86& 90& \textbf{91}\\
        RTE & acc & dev & 45.8& 58.8& 60.3& \textbf{67.9}& 51.3& 49.1& 63.8& 50.5& 63.5\\
        WiC & acc & dev & 50.0& 49.8& 49.5& 50.3& \textbf{50.8} & 50.3& 44& 50.6 & 0.0\\
        Multirc & f1a & dev & 57.7&	58.0&	52.4&	\textbf{73.7}&	58.6&	53.0&	39.0&	54.8&	72.9\\
        WSC & acc & dev & 65.6& 79.3& 81.8& \textbf{85.3}& 66.3& 77.2& 80.7& 82.8& 65.4\\
        ReCoRD & acc & dev & 77.5& 87.1& 88.9&	\textbf{90.3}&	71.6&	86.7&	89.2&\textbf{90.3}&	90.2\\
        CB & acc & dev & \textbf{66.1} & 33.9& 40.7& 48.2& 42.9& 37.5& 33.9& 42.9& 46.4\\
        \addlinespace
        ANLI R1 & acc & dev & 34.1& 33.9& 33.4& 39.2 & 36.1& 33.2& 34.7& \textbf{39.4} & 34.6\\
        ANLI R2 & acc & dev & 33.8& 32.4& 34.9& \textbf{37.3}& 36.7& 33.6& 34.8& 35.7& 35.4\\
        ANLI R3 & acc & dev & 32.8& 34.0& 34.6& \textbf{41.3}& 34.8& 34.1& 34.9& 34.6& 34.5\\
        \addlinespace
        Avg NLG & - & - & 18.6 & 35.1 & 39.6 & \textbf{54.6} & 14.9 & 31.3 & 38.0 & 45.8 & 47.6 \\ 
        Avg NLU & - & - & 51.5 & 58.3 & 61.1 & \textbf{66.2} & 48.9 & 56.1 & 60.2 & 63.2 & 60.8\\
        \bottomrule
    \end{tabular}

\end{table*}
\clearpage

\begin{table*}[htb]
    \centering
    \renewcommand\tabcolsep{5pt}
    \renewcommand{\arraystretch}{1.4}
            \caption{One-shot scores on all 29 benchmarks for GPT3 and different \glam MoE and dense models.}
        \label{tab:1shot}
        \vskip 0.1in
    \footnotesize
    \begin{tabular}{llcccccccccc}
        \toprule
      & & & \multicolumn{4}{c}{\bf \glam(MoE)} & \multicolumn{4}{c}{\bf \glam(Dense)} & {\bf GPT3}\\
        \cmidrule(lr){4-7} \cmidrule(lr){8-11} \cmidrule(lr){12-12}
        Name & Metric & Split & 0.1B/64E & 1.7B/64E & 8B/64E & 64B/64E & 0.1B & 1.7B & 8B & 137B & GPT-3 (175B)\\
        \midrule
        TriviaQA & acc (em) & dev & 15.2 & 54.1 & 65.9 & \textbf{75.8} & 8.3 & 36.3 & 56.4 & 70.0 & 68.0\\
        NQs & acc (em) & test & 2.5 & 10.7 & 16.0 & \textbf{26.3} & 1.19 & 6.5 & 10.7 & 19.1 & 23.0\\
        WebQS & acc (em) & test & 5.9 & 13.9 & 17.0 & 24.4 & 3.44 & 9.3 & 11.6 & 18.8 & \textbf{25.3}\\
        \addlinespace
        Lambada & acc (em) & test & 36.9 & 57.4 & 64.1 & \textbf{80.9} & 21.8 & 52.3 & 64.7 & 68.5 & 72.5\\
        HellaSwag & acc & dev & 43.5 & 66.4 & 74.0 & 76.8 & 34.7 & 60.5 & 72.6 & 76.8 & \textbf{78.1}\\
        StoryCloze & acc & test & 67.0 & 77.9 & 80.0 & 84.0 & 63.7 & 76.4 & 82.1 & 82.6 & \textbf{84.7}\\
        \addlinespace
        Winograd & acc & test & 69.2 & 80.2 & 85.3 & 83.9 & 65.6 & 80.2 & 84 & 85.3 & \textbf{89.7}\\
        WinoGrande & acc & dev & 51.7 & 63.5 & 68.7 & 73.0 & 49.8 & 62.8 & 70.0 & 73.1 & \textbf{73.2}\\
        \addlinespace
        DROP & f1 & dev & 16.3 & 24.8 & 28.4 & \textbf{57.8} & 19.3 & 24.9 & 41.2 & 49.4 & 34.3\\
        CoQA & f1 & dev& 48.3 & 72.8 & 76 & 79.6 & 33.3 & 72.7 & 74.4 & 78.8 & \textbf{84.0}\\
        QuAC & f1 & dev & 28.7 & 35.2 & 43.1 & 42.7 & 23.7 & 35.7 & 35.1 & \textbf{44.6} & 43.4\\
        SQuADv2 & f1 & dev & 35.5 & 69.5 & 76.3 & \textbf{71.8} & 34.2 & 67.1 & 69.2 & 70.0 & 65.4\\        
        SQuADv2 & acc (em) & dev & 21.8 & 53.6 & 60.9 & \textbf{66.5} & 29.0 & 50.8 & 64.2 & 63.7 & 60.1\\
        RACE-m & acc & test & 42.7 & 60.9 & 60.6 & 65.5 & 43.1 & 56.4 & 63.1 & \textbf{69.0} & 57.4\\
        RACE-h & acc & test & 29.1 & 41.9 & 44.6 & \textbf{48.7} & 29.4 & 40.8 & 45.3 & 47.7 & 45.9\\
        \addlinespace
        PIQA & acc & dev & 69.0 & 76.0 & 78.1 & \textbf{81.4} & 63.7 & 73.1 & 76.3 & 79.5 & 80.5\\
        ARC-e & acc & test & 53.5 & 68.1 & 73.4 & 76.6 & 45.9 & 63.8 & 62.6 & \textbf{77.2} & 71.2\\
        ARC-c & acc & test & 27.0 & 39.3 & 44.8 & 50.3 & 24.5 & 35.2 & 41.5 & 50.7 & \textbf{53.2}\\
        Openbookqa & acc & test & 39.6 & 47.6 & 50.6 & 55.2 & 37.8 & 47.2 & 53.0 & 55.4 & \textbf{58.8}\\
        \addlinespace
        BoolQ & acc & dev & 53.6 & 62.0 & 70.8 & \textbf{82.8} & 55.7 & 58.1 & 76.4 & 77.5 & 76.7\\
        Copa & acc & dev & 75 & 81 & 86 & \textbf{92} & 71 & 81 & 86 & 91 & 87\\
        RTE & acc & dev & 53.1 & 54.5 & 57.0 & \textbf{71.5} & 53.4 & 55.2 & 62.0 & 58.4 & 70.4\\
        WiC & acc & dev & 47.3 & 47.0 & 48.0 & \textbf{52.7} & 47.3 & 46.8 & 48.0 & 48.7 & 48.6\\
        Multirc & f1a & dev & 58.5 & 59.6 & 62.0 & \textbf{74.7} & 56.3 & 59.4 & 61.9 & 64.2 & 72.9\\
        WSC & acc & dev & 67.7 & 77.5 & 83.8 & 83.9 & 63.8 & 78.5 & 83.0 & \textbf{86.3} & 69.2\\
        ReCoRD & acc & dev & 77.5 & 87.3 & 89.0 & \textbf{90.3} & 71.6 & 86.2 & 89.2 & 90.2 & 90.1\\
        CB & acc & dev & 41.1 & 35.7 & 44.6 & \textbf{73.2} & 42.9 & 41.1 & 30.4 & 48.2 & 64.3\\
        \addlinespace
        ANLI R1 & acc & dev & 32.1 & 31.1 & 32.3 & \textbf{42.4} & 32.5 & 31.4 & 31.9 & 34.8 & 32.0\\
        ANLI R2 & acc & dev & 31.1 & 30.7 & 32.5 & \textbf{40.0} & 30.7 & 31.2 & 30.7 & 32.6 & 33.9\\
        ANLI R3 & acc & dev & 30.5 & 31.6 & 34.8 & \textbf{40.8} & 30.9 & 30.3 & 32.4 & 35.0 & 35.1\\
        \addlinespace
        Avg NLG & - & - & 23.5 & 43.6 & 49.7 & \textbf{58.4} & 19.4 & 39.5 & 47.5 & 52.8 & 52.7 \\ 
        Avg NLU & - & - & 50.4 & 58.1 & 61.9 & \textbf{68.6} & 48.3 & 56.9 & 61.7 & 65.0 & 65.4\\
        \bottomrule
    \end{tabular}
\end{table*}

\begin{table*}[htb]
    \centering
    \renewcommand\tabcolsep{5pt}
    \renewcommand{\arraystretch}{1.4}
            \caption{Few-shot scores on all 29 benchmarks for GPT3 and different \glam MoE and dense models. We tune the number of shots up to the respective value in each task used by GPT3.}
        \label{tab:kshot}
        \vskip 0.1in
    \footnotesize
    \begin{tabular}{llcccccccccc}
        \toprule
      & & & \multicolumn{4}{c}{\bf \glam(MoE)} & \multicolumn{4}{c}{\bf \glam(Dense)} & {\bf GPT3}\\
        \cmidrule(lr){4-7} \cmidrule(lr){8-11} \cmidrule(lr){12-12}
        Name & Metric & Split & 0.1B/64E & 1.7B/64E & 8B/64E & 64B/64E & 0.1B & 1.7B & 8B & 137B & GPT-3 (175B)\\
        \midrule
        TriviaQA & acc (em) & dev & 21.7 & 60.1 & 67.7 & \textbf{75.8} & 8.3 & 38.8 & 56.4 & 70.0 & 71.2\\
        NQs & acc (em) & test & 5.3 & 17.7 & 24.4 & \textbf{32.5} & 1.50 & 9.0 & 20.1 & 27.9 & 29.9\\
        WebQS & acc (em) & test & 12.1 & 24.4 & 29.6 & 41.1 & 6.90 & 9.3 & 25.5 & 32.9 & \textbf{41.5}\\
        \addlinespace
        Lambada & acc (em) & test & 36.9 & 64.3 & 79.0 & \textbf{86.6} & 21.8 & 63.0 & 77.1 & 84.2 & 86.4\\
        HellaSwag & acc & dev & 45.6 & 66.2 & 74.0 & 77.2 & 34.7 & 60.7 & 72.6 & 76.8 & \textbf{79.3}\\
        StoryCloze & acc & test & 69.4 & 80.0 & 82.8 & 86.7 & 63.7 & 78.7 & 83.7 & 85.7 & \textbf{87.7}\\
        \addlinespace
        Winograd & acc & test & 69.2 & 82.8 & 85.3 & \textbf{88.6} & 65.6 & 80.5 & 85.4 & 85.3 & \textbf{88.6}\\
        WinoGrande & acc & dev & 52.6 & 66.2 & 71.4 & \textbf{79.2} & 49.8 & 64.2 & 72.3 & 76.6 & 77.7\\
        \addlinespace
        DROP & f1 & dev & 23.5 & 37.0 & 40.0 & \textbf{58.6} & 19.3 & 41.4 & 49.4 & 49.4 & 36.5\\
        CoQA & f1 & dev& 48.3 & 66.0 & 72 & 79.6 & 33.3 & 66.0 & 74.4 & 78.8 & \textbf{85.0}\\
        QuAC & f1 & dev & 26.0 & 34.2 & 43.1 & 42.8 & 23.7 & 34.3 & 35.1 & 37.2 & \textbf{44.3}\\
        SQuADv2 & f1 & dev & 38.7 & 61.8 & 67.1 & \textbf{71.8} & 34.2 & 60.0 & 69.6 & 70.0 & 69.8\\        
        SQuADv2 & acc (em) & dev & 32.7 & 55.5 & 60.9 & \textbf{67.0} & 29.0 & 53.9 & 64.2 & 63.7 & 64.9\\
        RACE-m & acc & test & 41.8 & 53.6 & 60.6 & \textbf{66.9} & 43.1 & 56.5 & 56 & 65.1 & 58.1\\
        RACE-h & acc & test & 31.5 & 40.2 & 44.6 & \textbf{49.3} & 29.5 & 40.8 & 43 & 48.1 & 46.8\\
        \addlinespace
        PIQA & acc & dev & 69.0 & 76.1 & 78.1 & 81.8 & 64.2 & 73.1 & 77 & 80.8 & \textbf{82.3}\\
        ARC-e & acc & test & 57.8 & 70.1 & 75.3 & \textbf{78.9} & 48.9 & 66.0 & 74 & 79.0 & 70.1\\
        ARC-c & acc & test & 29.7 & 38.3 & 45.5 & 52.0 & 24.8 & 35.2 & 41.5 & 45.7 & 51.5\\
        Openbookqa & acc & test & 41.6 & 49.6 & 53.0 & 63.0 & 37.8 & 54 & 54.0 & 58.8 & \textbf{65.4}\\
        \addlinespace
        BoolQ & acc & dev & 53.6 & 62.0 & 70.5 & \textbf{83.1} & 59.9 & 63.1 & 76.4 & 80.5 & 77.5\\
        Copa & acc & dev & 75 & 82 & 88 & \textbf{93.0} & 71 & 83 & 92.0 & 91.0 & 92.0\\
        RTE & acc & dev & 53.1 & 54.5 & 60.0 & \textbf{76.2} & 54.9 & 55.2 & 64.0 & 63.9 & 72.9\\
        WiC & acc & dev & 49.4 & 51.3 & 53.3 & \textbf{56.3} & 51.9 & 50.9 & 50.0 & 53.6 & 55.3\\
        Multirc & f1a & dev & 58.5 & 59.7 & 62.0 & \textbf{77.5} & 56.3 & 59.4 & 61.5 & 68.1 & 74.8\\
        WSC & acc & dev & 67.7 & 80.4 & 83.8 & 85.6 & 65.6 & 80.0 & 82.0 & \textbf{87.4} & 75.0\\
        ReCoRD & acc & dev & 77.5 & 87.3 & 89.0 & \textbf{90.6} & 71.8 & 86.2 & 89.0 & 90.5 & 89.0\\
        CB & acc & dev & 43.0 & 53.6 & 60.7 & \textbf{84.0} & 42.9 & 55.4 & 58 & 53.6 & 82.1\\
        \addlinespace
        ANLI R1 & acc & dev & 34.3 & 31.4 & 34.0 & \textbf{44.3} & 33.5 & 33.1 & 33.2 & 35.8 & 36.8\\
        ANLI R2 & acc & dev & 32.3 & 33.0 & 32.0 & \textbf{41.2} & 34.4 & 33.7 & 33.9 & 35.6 & 34.0\\
        ANLI R3 & acc & dev & 33.9 & 35.8 & 33.0 & \textbf{44.7} & 32.9 & 33.3 & 35.0 & 34.7 & 40.2\\
        \addlinespace
        Avg NLG & - & - & 27.2 & 46.8 & 53.0 & \textbf{61.6} & 19.8 & 42.7 & 52.4 & 57.1 & 58.8 \\ 
        Avg NLU & - & - & 51.7 & 59.7 & 63.6 & \textbf{71.4} & 49.2 & 59.2 & 63.7 & 66.8 & 68.4\\
        \bottomrule
    \end{tabular}
\end{table*}

\end{document}